\documentclass{article}
\usepackage[nonatbib, preprint]{neurips_2024}
\usepackage[utf8]{inputenc} 
\usepackage[T1]{fontenc}    
\usepackage{hyperref}       
\usepackage{url}            
\usepackage{amsfonts}       
\usepackage{nicefrac}       
\usepackage{microtype}      
\usepackage{xcolor}         
\usepackage{subcaption}
\usepackage{graphicx}
\usepackage{parskip}
\usepackage{booktabs}
\usepackage{amsmath}
\usepackage{multicol}

\begin{document}

\title{
    CNNs for Style Transfer of Digital to Film Photography 
}

\author{
  Pierre Mackenzie\thanks{\url{https://pierre.wiki}} \\
  EPFL\\
  \texttt{p.r.lardet@sms.ed.ac.uk} \\
  \And
    Mika Senghaas\thanks{\url{https://www.mikasenghaas.de/}} \\
    EPFL\\
    \And
    Raphaël Achddou \\
    EPFL\\
}

\maketitle

\begin{abstract}

The use of deep learning in stylistic effect generation has seen increasing use over recent years. In this work, we use simple convolutional neural networks to model Cinestill800T film given a digital input. We test the effect of different loss functions, the addition of an input noise channel and the use of random scales of patches during training. We find that a combination of MSE/VGG loss gives the best colour production and that some grain can be produced, but it is not of a high quality, and no halation is produced. We contribute our dataset of aligned paired images taken with a film and digital camera for further work.

\end{abstract}

\section{Introduction}
\label{sec:introduction}

Film photography is popular among artists and photographers for capturing scenes
in a unique way. They use optical flaws, such as halation and chromatic
aberrations, for aesthetic purposes. Such visual effects are the result of
complex physical processes that are not present in modern digital photography.
Hence, to recreate the visual effects of film in digital images, computational
methods are required. This motivates our work, where we investigate the
feasibility of using a pure deep learning approach to recreate the visual
effects of Cinestill-800T film, an iconic film stock known for its grain
textures, color hues and red halation.

Most available solutions involve directly modelling
the physical processes that lead to the visual effects~\cite{film-grain-rendering}. However, the complex
interplay between the camera, film stock, and chemicals during the development
process, means that simplifying assumptions have to be made. In addition, such
models are often computationally expensive and time consuming, especially when applied to
high-resolution images.

Ideally, a digital image could be processed into its film equivalent on the
order of seconds with high fidelity. Deep learning models have shown impressive
results in learning complex mappings, out-performing handcrafted models in many
tasks. One can hope that a deep learning model could learn higher quality
style translations than existing methods, while reducing computational complexity.
However, the problem of recreating film effect has not been directly addressed
by deep learning literature. Related computational photography problems include
camera raw-to-raw mappings, quality enhancement, grayscale colorisation, and
style transfer. As none of these involve direct applicatino to the problem of film
effect generation, we examine the commonly used techniques and how they could be adapted to our task.

All of this motivates our work, where we investigate the feasibility of using
deep learning to recreate the visual effects of Cinestill-800T film in digital
images. We propose training a U-Net-like model translation network on a
new dataset of paired digital and film images. While we acknowledge
that using direct statistical models for film grain, halation, and color hue may
be effective as a standalone, or in combination with deep learning, we constrain
this work to only explore pure deep learning methods.

Our contributions can be summarised as:

\begin{itemize}
    \item We propose a pure deep learning approach to recreate the visual effects of Cinestill-800T film in digital images.
    \item We present a self-collected dataset of paired digital and film images, which we publicly release to the research community.
    \item We investigate and analyse various techniques that help to improve the creation of visual effects, such as adding noise to the input, training on random-resized crops, or varying the loss function.
\end{itemize}
\section{Literature Review}
\label{sec:literature-review}

Many methods have presented as viable solutions for stylistic effect generation using image to image translation using Deep Learning. Among these, there is a variety of kinds of model architectures and effects produced.

\vspace{5pt}\noindent\textbf{Image Translation.} GANs have seen successful application in generic image translation, such as in Conditional Image-Image Translation \cite{conditionalgan} in which a loss is learned by a discriminator network which is conditioned on the input. This can be generalised to unpaired image data using Cyclic GANs \cite{cyclicgan} where two translation networks are used in either direction, and cyclic consistency is enforced using the loss function.

\noindent\textbf{Super Resolution.} The task of generating a higher resolution image from a lower resolution image has been tackled by deep learning approaches with a variety of pixel-wise and perceptual losses \cite{image-super-resolution-with-deep-networks,accurate-image-super-resolution,resolution-perceptual-losses}. GANs have produced impressive results \cite{gan-photo-realistic-super-resolution, esrgan-super-resolution}, especially in combination with perceptual losses \cite{pulse-gan-perceptual-super-resolution,recovering-texture-super-resolution}. Methods for greater stability during training have been developed \cite{gan-progressive-stability-resolution} and for greater control over which style is produced \cite{gan-style-based-resolution}. 

\noindent A contextual bilateral loss was introduced by \cite{zoom-to-learn} based on RGB and VGG feature patches that builds on the contextual loss proposed by \cite{contextual-loss} by adding a distance weighting. The authors claim that this loss is robust to misalignment in paired datasets. We implemented this loss but found it did not improve our results and do not include it in our final models.

\noindent\textbf{Image Colorisation.} Converting a grayscale image to a color has been attempted by \cite{large-scale-colorisation} using chromaticity maps, and \cite{intrinsic-colorisation} addresses the problem of illuminants by searching the web for reference images. \cite{deep-colorisation} leverages deep learning with separate networks for different kinds of images. This suggests a natural extension of our work would be to condition the digital-film mapping on features such as a luminance histogram.

\noindent\textbf{Raw to Raw Mapping.} \cite{raw-to-raw} maps the raw output of one camera to another using paired auto-encoders. Each auto-encoder is trained on raw outputs from a different camera, and the latent space of the two auto-encoders is mapped using a paired dataset. We implemented this architecture for a digital to film mapping, but it did not produce good initial results so we did not pursue it further.

\noindent\textbf{Style Transfer.} Style transfer, which maps the style of one image to the content of another, has seen success using CNNs which are trained to minimize a content and style loss. This has been applied to paintings \cite{image-style-transfer, scaling-painting-style-transfer} and to photorealistic images \cite{deep-photo-style-transfer}. These methods require an expensive optimization for every image. Instead, \cite{resolution-perceptual-losses} uses perceptual losses based on VGG features to train a feed forward model and \cite{arbitrary-style-transfer} extends this to arbitrary styles using adaptive instance normalization. \cite{dslr-quality} attempts to convert the style of phone images to DSLR quality using a composite loss including functions for colour, texture and content. We follow a similar approach in our work, using perceptual features produced by VGG-19 \cite{vgg} as part of our loss to train a feed forward CNN for style transfer.
\section{Implementation}
\label{sec:implementation}

To recreate the look of Cinestill-800T we aim to train a deep image-to-image
translation network on a paired digital-film dataset. This requires a dataset of
digital images and their corresponding film images, a suitable model
architecture, and loss functions capable of capturing the appearance of film.

\subsection{Data Collection and Preprocessing}
\label{subsec:data}
\label{subsubsec:data-collection}

While unpaired digital or film datasets are abundant, our method requires paired digital-film images. This means that the exact same scene has to be captured with both a digital and a film camera to allow the model to learn the differences in appearance between the two media. Images should be aligned, as our model should only transfer style and not learn move contents of the image around, and diverse in order to capture Cinestill-800T's baheviour in a variety of scenes and lighting conditions. As no such dataset was available, we collected our own.

\begin{figure}
    \begin{subfigure}[t]{.19\textwidth}
      \centering
      \includegraphics[width=\linewidth]{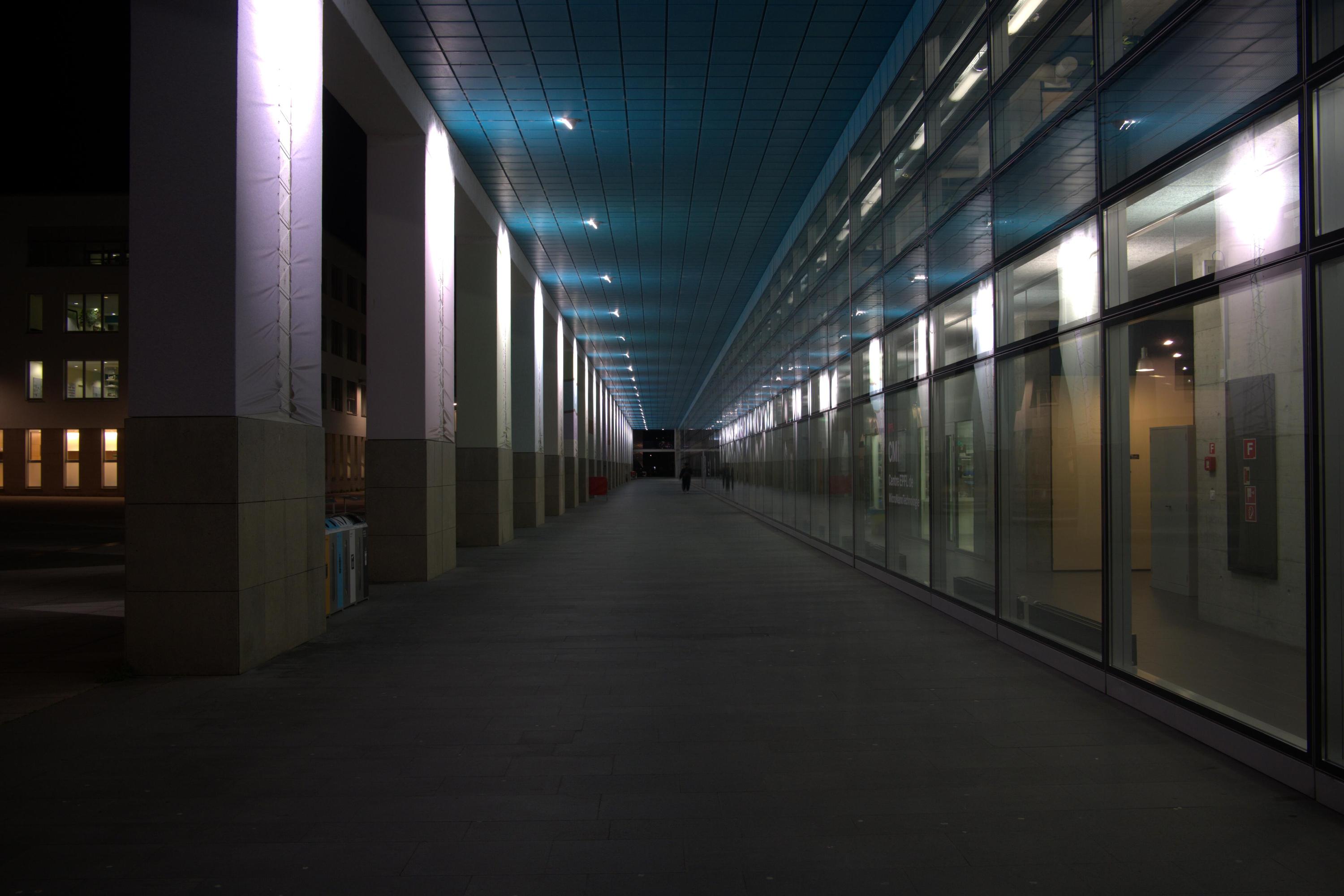}
    \end{subfigure}
    \hfill
    \begin{subfigure}[t]{.19\textwidth}
      \centering
      \includegraphics[width=\linewidth]{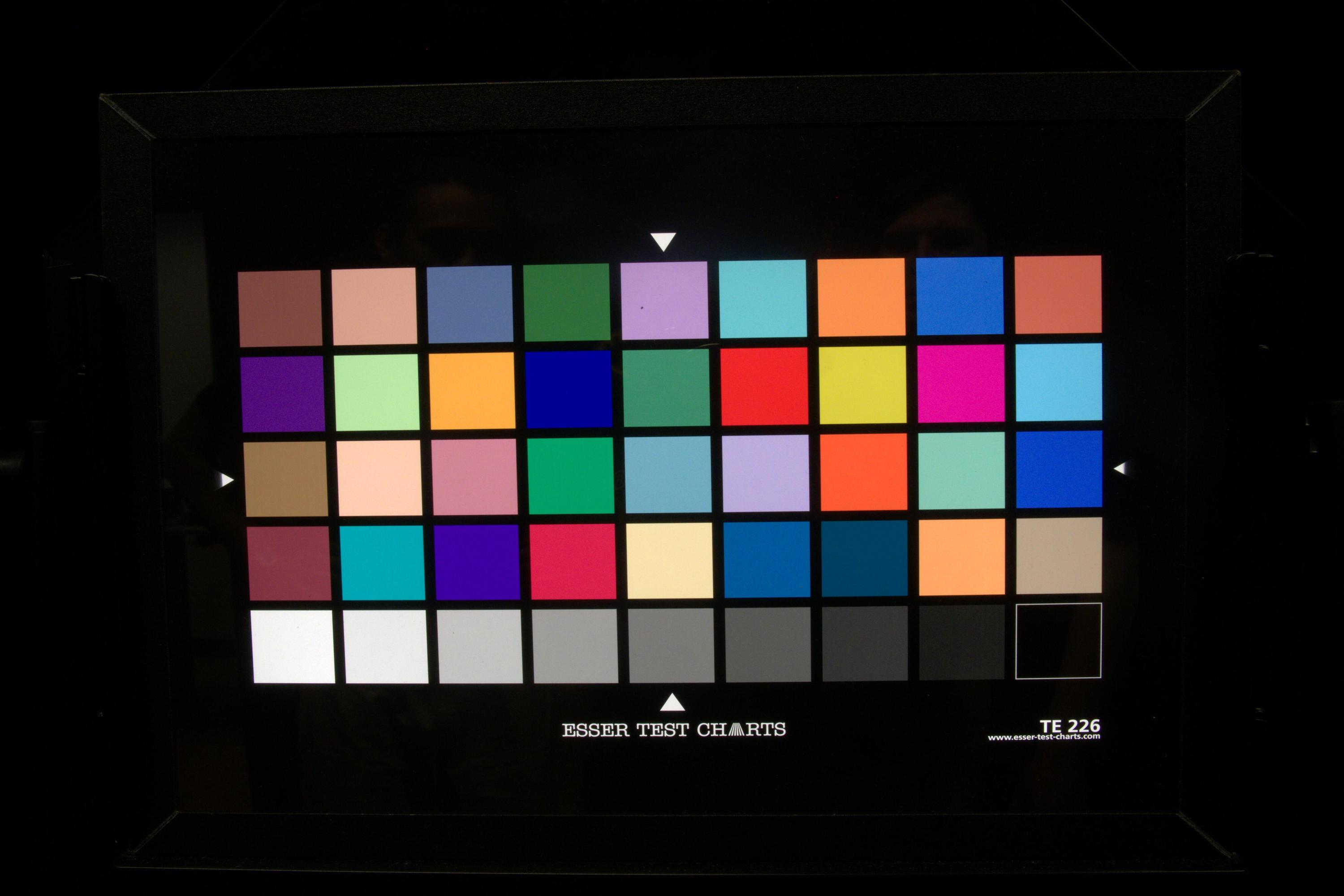}
    \end{subfigure}
    \hfill
    \begin{subfigure}[t]{.19\textwidth}
      \centering
      \includegraphics[width=\linewidth]{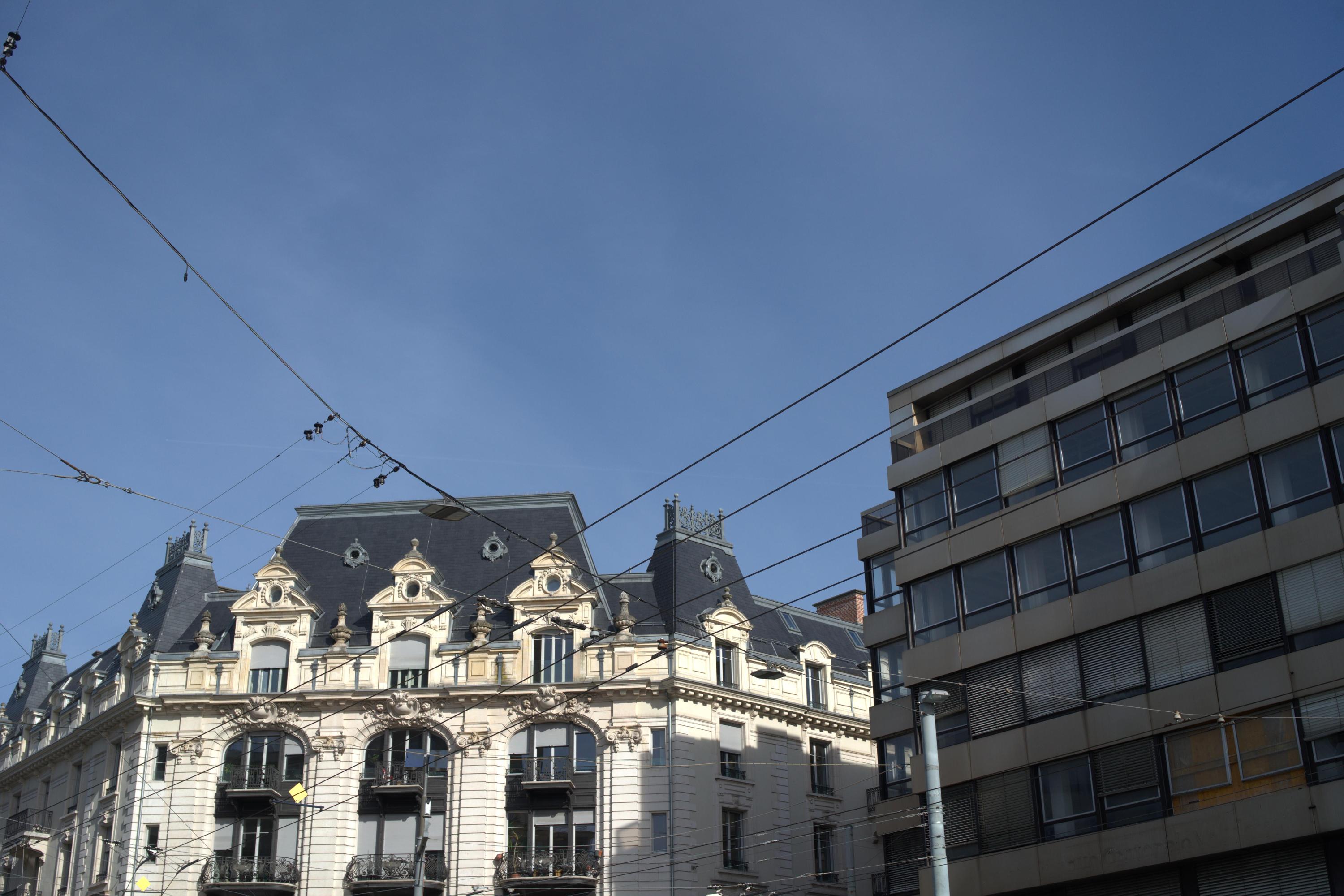}
    \end{subfigure}
    \hfill
    \begin{subfigure}[t]{.19\textwidth}
      \centering
      \includegraphics[width=\linewidth]{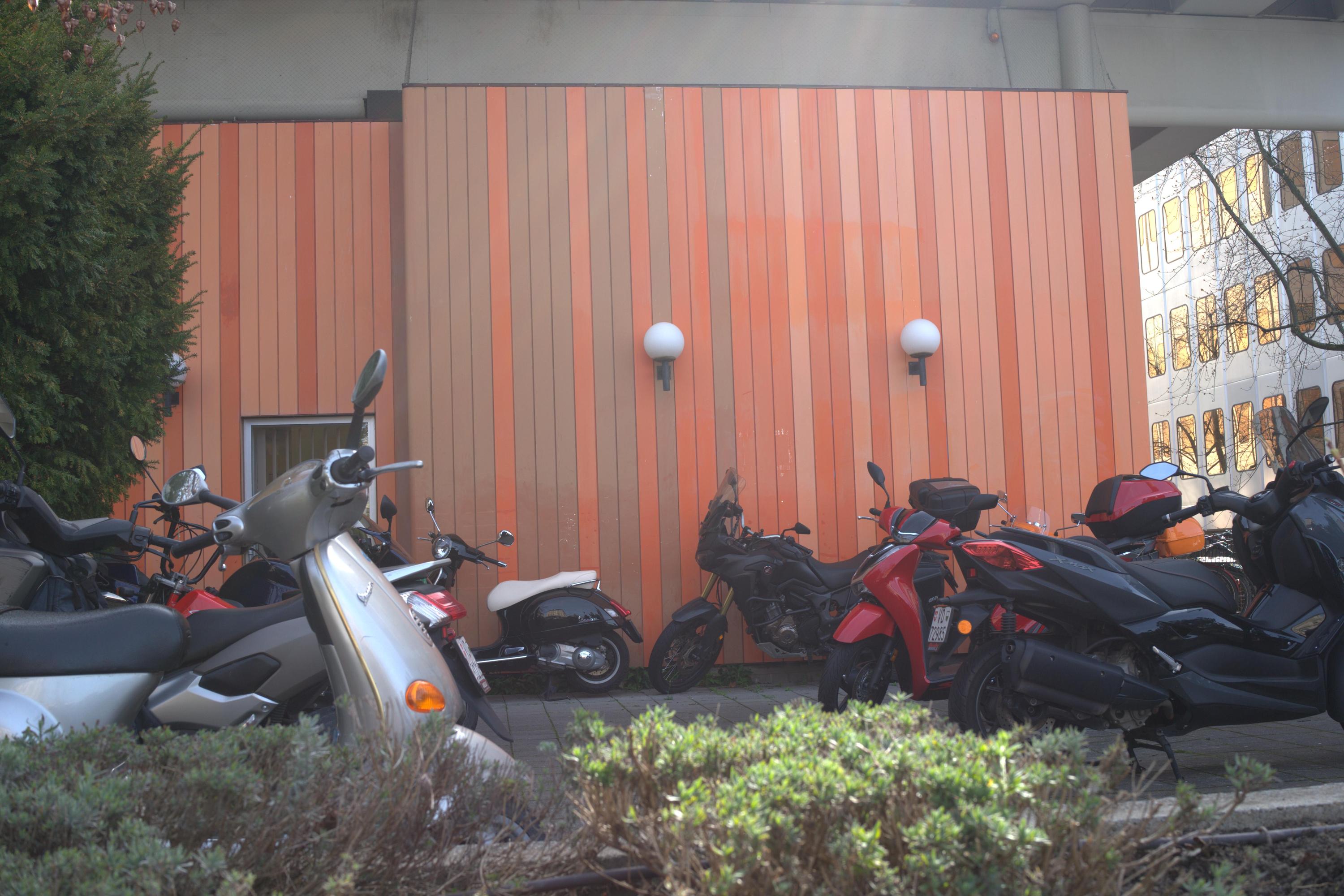}
    \end{subfigure}
    \begin{subfigure}[t]{.19\textwidth}
      \centering
      \includegraphics[width=\linewidth]{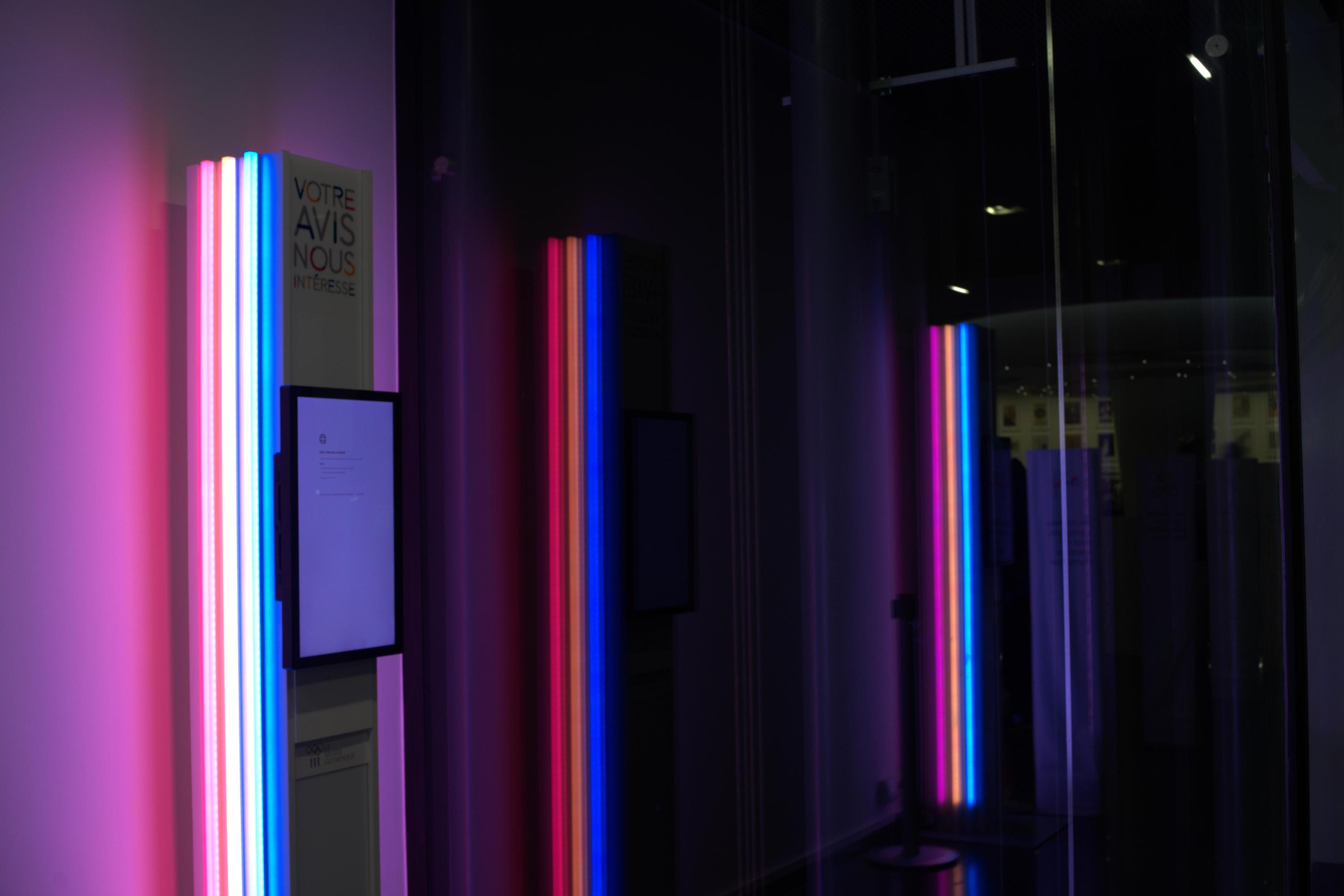}
    \end{subfigure}
  
    \medskip

    \begin{subfigure}[t]{.19\textwidth}
      \centering
      \includegraphics[width=\linewidth]{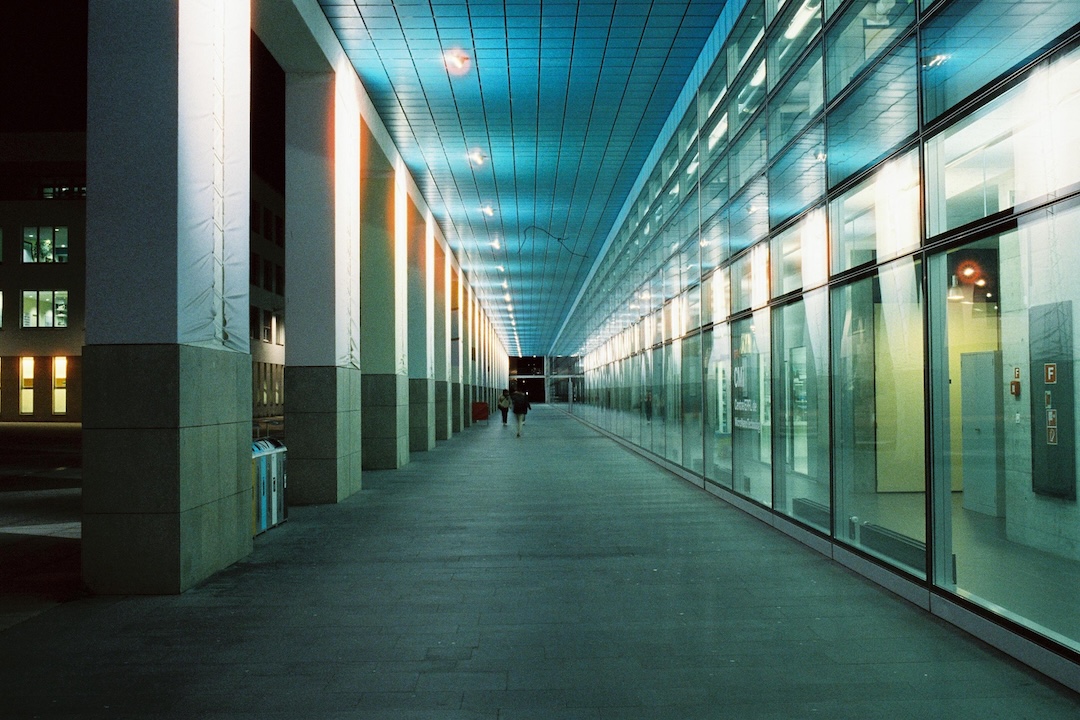}
    \end{subfigure}
    \hfill
    \begin{subfigure}[t]{.19\textwidth}
      \centering
      \includegraphics[width=\linewidth]{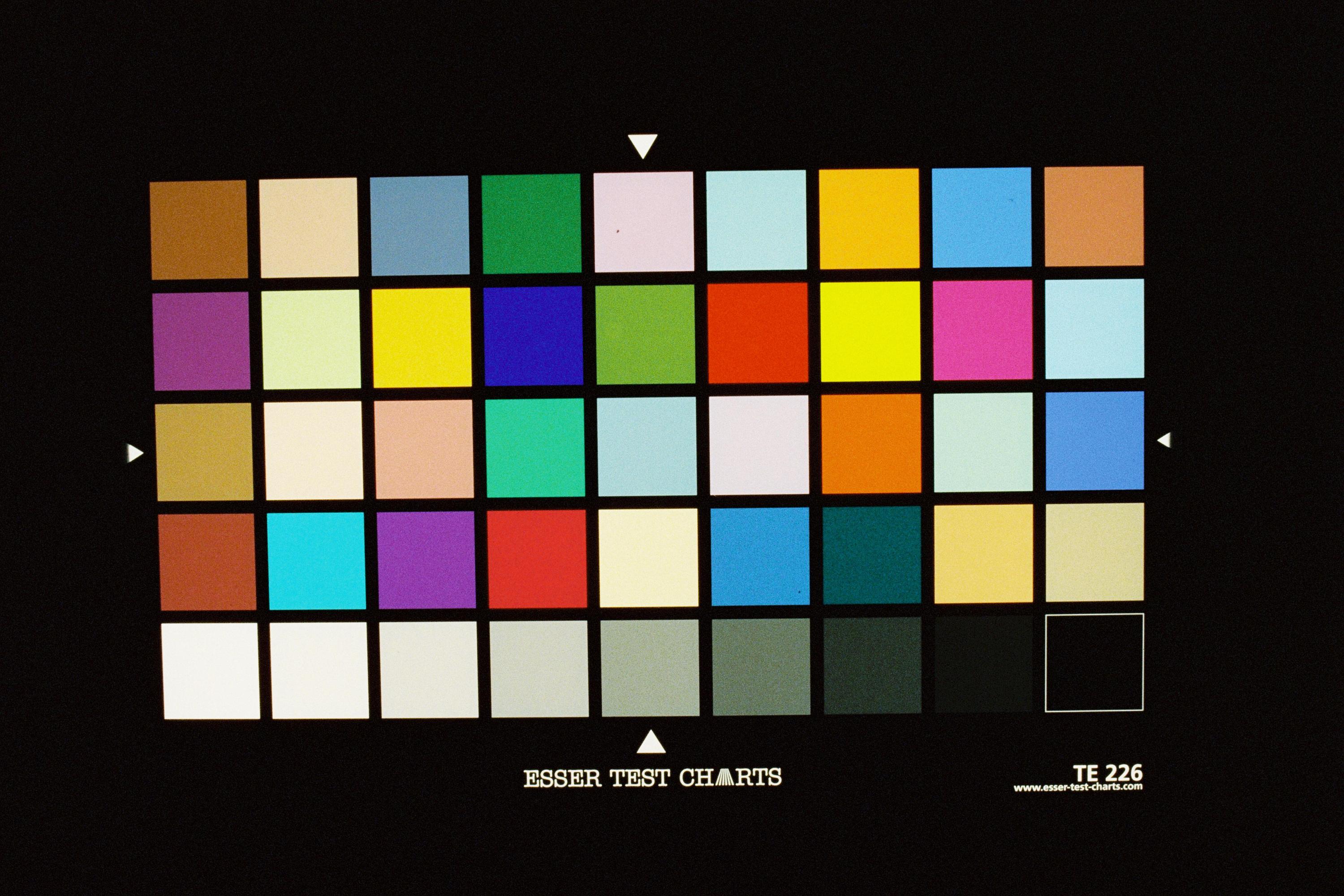}
    \end{subfigure}
    \hfill
    \begin{subfigure}[t]{.19\textwidth}
      \centering
      \includegraphics[width=\linewidth]{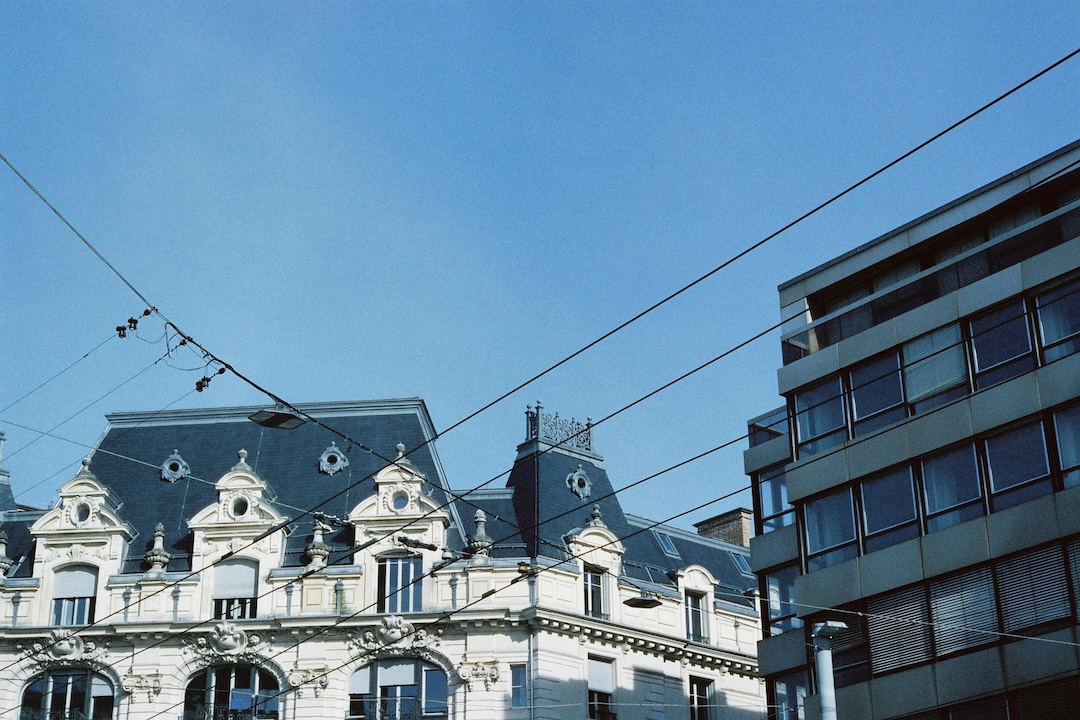}
    \end{subfigure}
    \hfill
    \begin{subfigure}[t]{.19\textwidth}
      \centering
      \includegraphics[width=\linewidth]{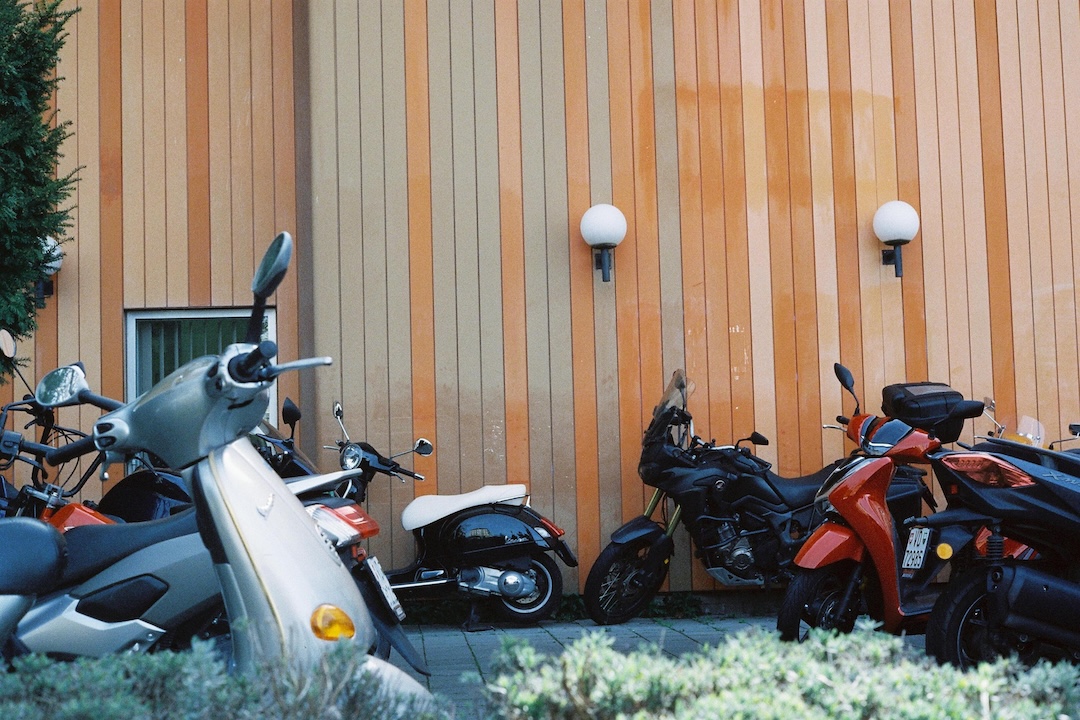}
    \end{subfigure}
    \begin{subfigure}[t]{.19\textwidth}
      \centering
      \includegraphics[width=\linewidth]{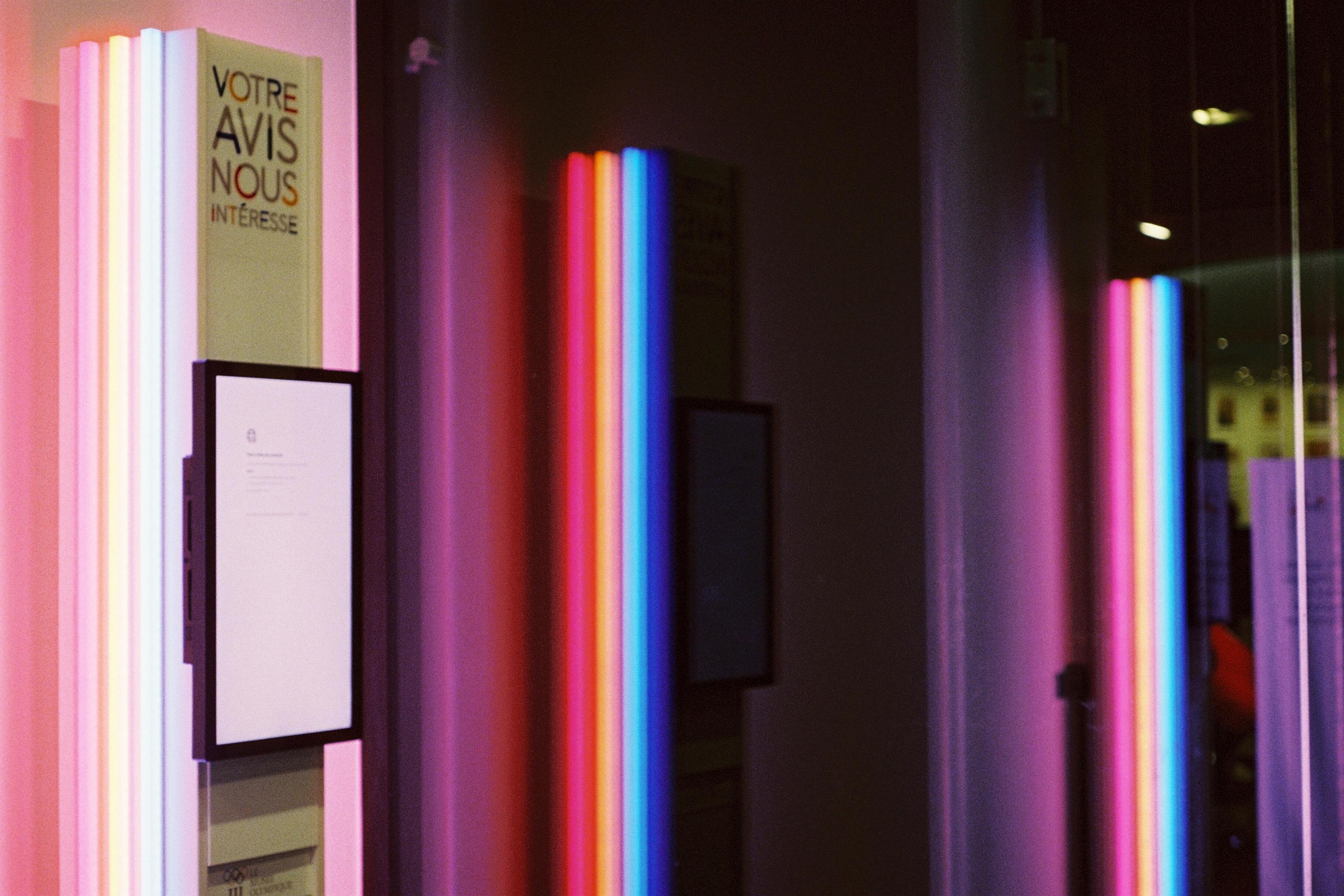}
    \end{subfigure}
  
    \caption{\textbf{Raw Paired Image Dataset.} Examples of raw image pairs from the dataset. A column shows a single scene captured with a digital camera (top) and a film camera (bottom). The images show a wide range of scenes and visual effects.}
    \label{fig:raw-dataset}
\end{figure}

\textbf{Data Collection.} To ensure aligned images, we take inspiration from the setup proposed in~\cite{dslr-quality} and capture images with two cameras, a Sony Alpha 7 digital camera and a Nikon F3 film camera. Photos were taken by keeping the same tripod setup for both cameras for each image. Camera settings like aperture, ISO and shutter speed were kept constant across both cameras as best as possible (Table~\ref{tab:cameras}). We only captured static scenes to limit changes in the scene between shots. To create a diverse dataset, images were taken both indoors and outdoors, in different weather conditions, with different light and containing different subjects. Specific emphasis was put on ensuring that all special visual effects (halation, grain and colours) were well-represented in the dataset. The final raw dataset consists of 41 image pairs. Some examples are presented in Figure~\ref{fig:raw-dataset}. 

\begin{table}
    \centering
    \caption{Camera Setup.}
    \begin{tabular}{lccccc}
        \toprule
        \textbf{Camera} & \textbf{Resolution} & \textbf{Lens} & \textbf{Aperture} & \textbf{ISO} & \textbf{Shutter} \\
        \midrule
        Sony Alpha 7 (Digital) & 24 MP & 40mm & F8 & 100 & Auto \\
        Nikon F3 (Film) & 12 MP & 55mm & F8 & 100 & Auto\\
        \bottomrule
    \end{tabular}
    \label{tab:cameras}
\end{table}

\subsubsection*{Data Preprocessing}
\label{subsubsec:preprocessing}

Inevitably, raw image pairs are not perfectly aligned. To address this
issue, we first spatially align the images using keypoint alignment. We follow a standard processing pipeline using openly available implementations of common feature detection, matching and perspective transformation algorithms from the OpenCV library~\cite{opencv}. For all image pairs we first detect keypoints and extract descriptors using ORB~\cite{orb}, match them using FLANN~\cite{flann}, and estimate a homography transformation matrix using the \href{https://www.mathworks.com/discovery/ransac.html}{RANSAC} algorithm.

To ensure that our model does not learn systematic brightness adjustments, we also align the luminance of each image pair. We align the luminance of the film image to the digital image as our model receives digital as input at inference time. The luminance alignment is performed using histogram matching on the luminance channel in CIELAB space before converting back to RGB.

\begin{figure}

    \begin{subfigure}[b]{0.49\textwidth}
      \centering
      \includegraphics[width=0.49\textwidth]{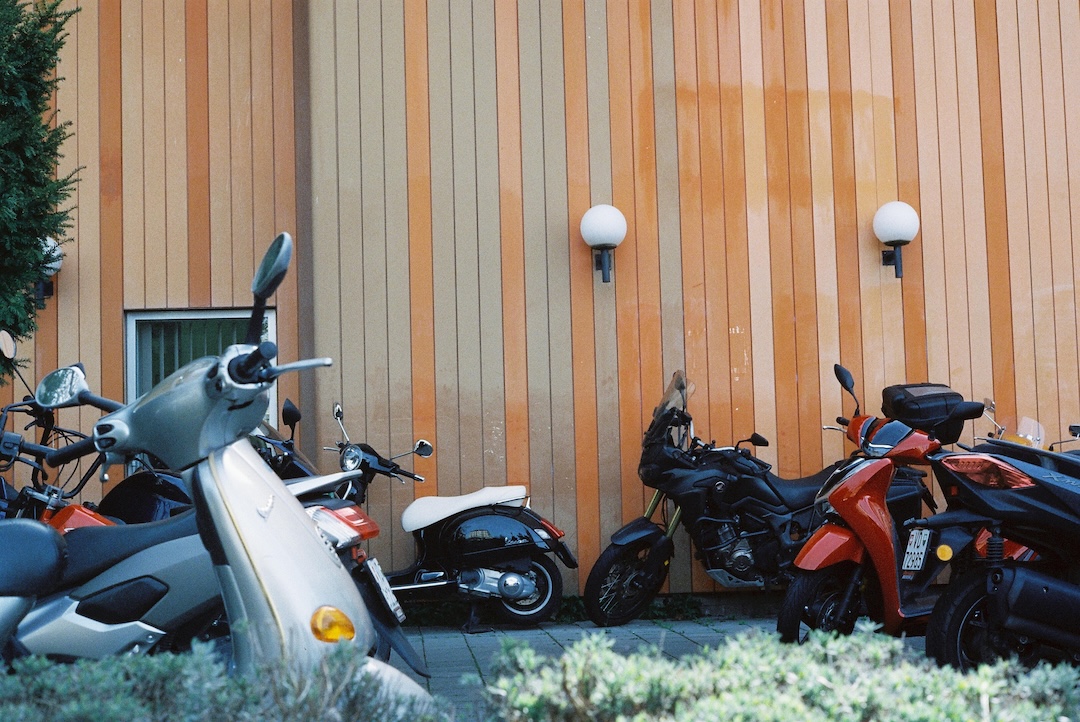}
      \includegraphics[width=0.49\textwidth]{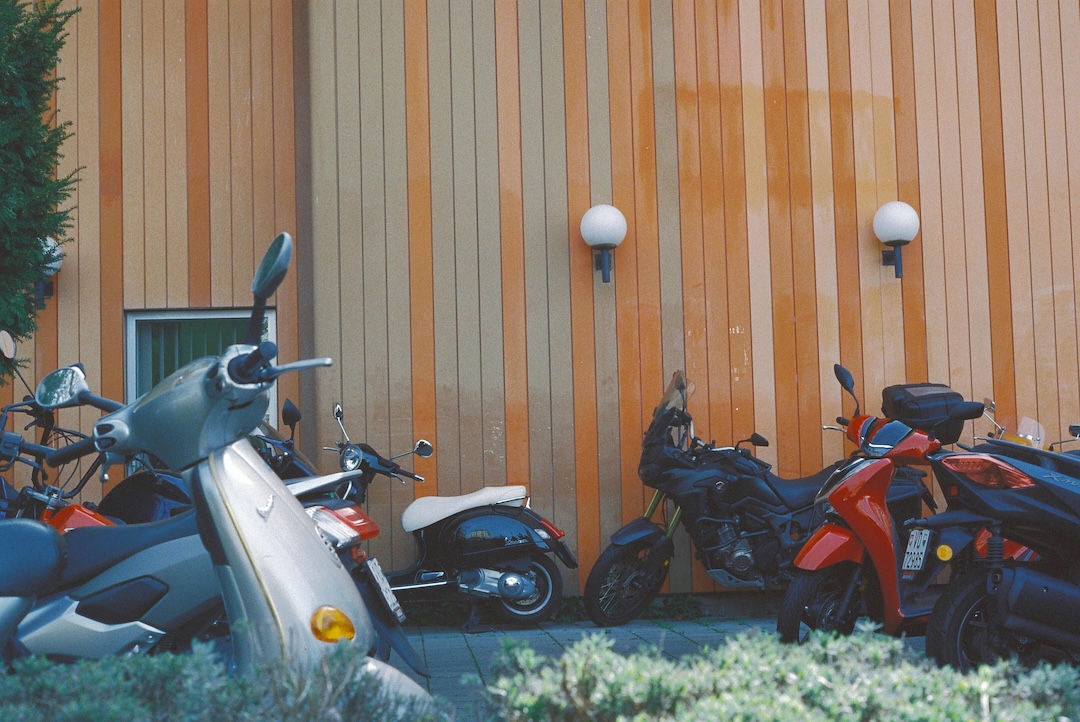}
      \captionsetup{justification=centering}
      \caption{Film: Raw (left) vs. Processed (right).} 
  \end{subfigure}
  \hfill
  \begin{subfigure}[b]{0.49\textwidth}
      \centering
      \includegraphics[width=0.49\textwidth]{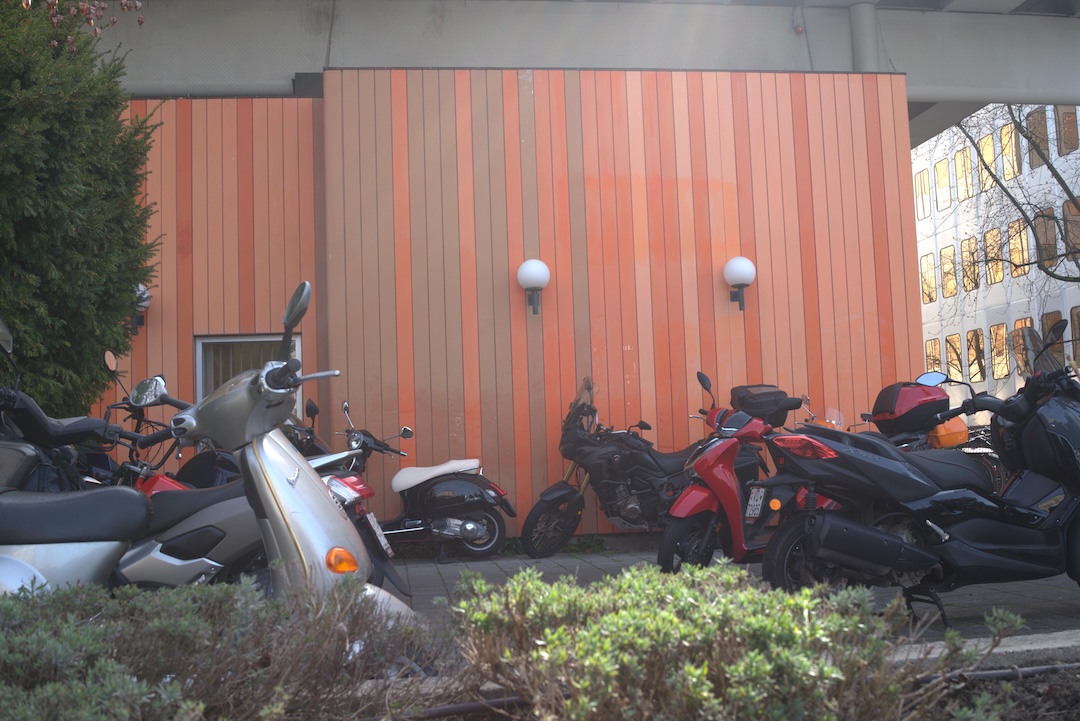}
      \includegraphics[width=0.49\textwidth]{raw-film}
      \captionsetup{justification=centering}
      \caption{Digital: Raw (left) vs. Processed (right).}
  \end{subfigure}

    \caption{\textbf{Dataset Preprocessing.} Example of a raw and processed image pair. We can see the luminance alignment in the film image (a) and the spatial alignment in the digital image (b).}
    \label{fig:data-preprocessing}
\end{figure}

An example of the results of these two steps can be seen in Figure~\ref{fig:data-preprocessing}. After the
preprocessing steps we are left with 38 processed image pairs. We publicly
release both the raw and processed dataset to facilitate further work.

\subsection{Model}
\label{subsec:model}

Our image style translation network is based on the U-Net
architecture~\cite{unet}. U-Net is a type of convolutional neural network (CNN)
originally introduced for image segmentation tasks in 2015. CNNs have been
successfully adapted for various image-to-image translation tasks and similar models are used as the
building block of many modern model families, such as auto-encoders \cite{raw-to-raw} and GANs \cite{perceptual-losses-style-transfer}. The network features a U-shaped architecture, composed of an encoder and a decoder. The encoder
resembles a conventional convolutional network, involving repeated convolutions
followed by a non-linear activation functions and pooling operations. The decoder integrates both feature and spatial information using up-convolutions and high-resolution features via skip connections.

We selected UNet as it is a relatively simple model family, allowing us to focus on
the specifics of our problem rather than the intricacies of the model. Despite its simplicity, similar models such as CNNs with residual connections have been shown to be effective for image-to-image translation, such as in \cite{dslr-quality}. As it is fully convolutional, it can be applied to images of arbitrary size.
\footnote{So long as the input dimensions are divisible by $2^L$, where $L$ is
the number of layers in the encoder}

For our experiments we follow related work \cite{unet,dslr-quality,raw-to-raw}
to set sensible hyperparameters for our model. Our image style translation
network is fully convolutional, with a three-layer encoder and decoder. The encoder consists of three convolutional blocks with 64, 128, and 256 filters, respectively. Each block consists of two convolutional layers with 3×3 kernels and zero padding, interleaved with ReLu activation functions.
In between each of the convolutional blocks, we downsample the spatial dimension
of the feature maps by a factor of two using a 2×2 max-pooling operation. The
decoder performs the reverse operation by upsampling the spatial
dimension. It decreases the number of filters from 256 to 128, 64, and finally
to 3. Upsampling is performed via a 2×2 transposed convolution. After
upsampling, we concatenate the feature maps from the corresponding layer in the
encoder before applying a convolutional block. After the final layer,
the model outputs a three-channel image, corresponding to the RGB channels of
the transformed image. Thus, the overall translation network $F_{\theta}$
is an image-to-image model, that takes an image $I$ as input and outputs a
transformed image $F_{\theta}(I)$ of the same dimensions. 
\subsection{Loss Functions}
\label{subsec:loss-functions}

We propose a list of loss functions that can be combined, each targeting different effects.

\textbf{MSE/MAE.} Both the mean squared error (MSE, Eq.~\ref{eq:mse}) and mean absolute error (MAE, Eq.~\ref{eq:mae}) are common pixel-to-pixel loss functions, that are popular in image reconstruction tasks. They measure the average squared or absolute difference between the predicted and ground truth images pixels, respectively. For two images, $X, Y \in \mathbb{R}^{3 \times H \times W}$, the MSE and MAE losses are defined as:
\begin{multicols}{2}
\begin{equation}
    \mathcal{L}_\text{MSE}(X, Y) = \frac{1}{n} \sum_{i=1}^{n} (X_i, Y_i)^2
    \label{eq:mse}
\end{equation}

\begin{equation}
    \mathcal{L}_\text{MAE}(X, Y) = \frac{1}{n} \sum_{i=1}^{n} |X_i - Y_i|
    \label{eq:mae}
\end{equation}
\end{multicols}

where $i$ indexes the pixels of the images. We hypothesise that
such loss functions are useful in preserving the overall structure and color of the image,
but are not sufficient to capture higher-level visual effects such as grain and
halation.

\textbf{VGG Loss.} VGG loss computes the MSE (Eq.~\ref{eq:mse}) between the feature representations of the predicted and ground truth images when passed through a pre-trained VGG-19~\cite{vgg} convolutional network. Letting $\varphi_k(I)$ denote the feature maps of the $k$-th layer of the VGG network, and the sets $K$ and $\lambda$ denote the layers of interest \footnote{We use conv1\_2, conv2\_2, conv3\_2 with weights 0.4, 0.4 and 0.2 respectively.} and their weights in the loss function respectively, the VGG loss is defined as:

\begin{equation}
   \mathcal{L}_\text{VGG}(X, Y) = \sum_{k \in K} \lambda_k \mathcal{L}_\text{MSE}(\varphi_k(X), \varphi_k(Y)).
    \label{eq:vgg}
\end{equation}

Intuitively, the VGG loss encourages the model to generate images that are similar to the ground truth images in terms of their high-level semantics such 
as edges, textures and patterns.

\textbf{Color Loss.} Inspired by~\cite{dslr-quality}, we consider an alternative to the MSE loss, which we refer to as the color loss. The color loss computes the MSE between two images, $X_b$ an $Y_b$, where $X_b$ and $Y_b$ are blurred versions of the predicted and ground truth images respectively. Blurring is done using a Gaussian filter with a kernel size of 7 and \(\sigma=3\). We can write:

\begin{equation}
    \mathcal{L}_\text{Color}(X, Y) = \mathcal{L}_\text{MSE}(X_b, Y_b).
    \label{eq:color}
\end{equation}

As the authors of the original paper argue, the main idea behind the loss is to evaluate the brightness, contrast and major colours of the image while ignoring fine-grained texture and content comparison. Crucially, the color loss, unlike MSE or MAE, is more robust to small pixel shifts which are present in our training data.

\textbf{Relative Total Variational Loss (TV-Rel).} Traditionally, the \href{https://lightning.ai/docs/torchmetrics/stable/image/total_variation.html}{total variational} loss is used to encourage image smoothness and reduce noise by minimising the sum of the absolute differences between shifted pixel values. In our case, we aim to create grain, which can be seen as a form of noise. To this end, we propose the relative total variational loss defined as the absolute difference between the total variation of the predicted and ground
truth images. 

\begin{equation}
    \mathcal{L}_\text{TV-Rel} = |TV(X) - TV(Y)| 
    \label{eq:tv-rel}
  \end{equation}

This loss encourages the model to generate images with a similar
amount of noise to the ground truth film images.

\textbf{Combination of Losses.} We combine the individual loss functions to form the final loss function using a weighted sum. For a set of loss functions $\mathcal{L}$ and corresponding weights $\beta$ \footnote{We simply use equal weights in all experiments with $\beta_i = 1$ for all $i$}, the final loss function is defined as

\begin{equation}
    \mathcal{L}(X, Y) = \sum_{i} \beta_i \mathcal{L}_i(X, Y).
    \label{eq:combined-loss}
\end{equation}

Some combinations of loss functions are more effective than others, and we will experiment with different combinations, as outlined in Section \ref{sec:results}. 
\subsection{Experiments}
\label{subsec:experiments}

\begin{figure}
  \centering
  \includegraphics[width=0.9\textwidth]{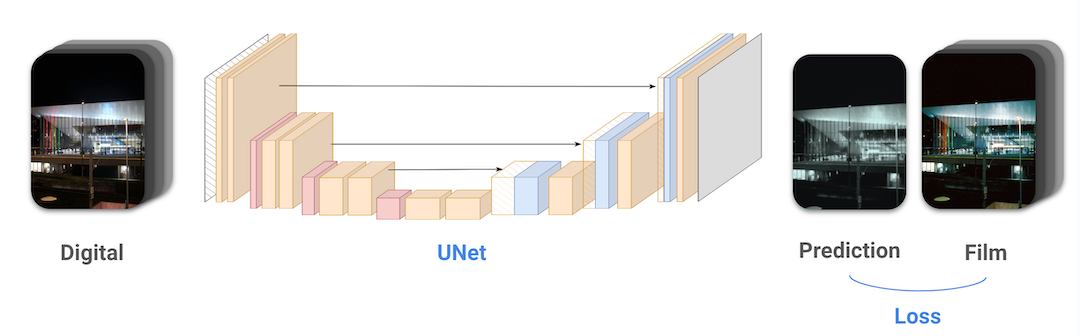}
  \caption{\textbf{Training Overview.} We train our model on paired digital-film images.}
  \label{fig:training-overview}
\end{figure}

The training procedure is illustrated in Figure~\ref{fig:training-overview}. We train our model on a dataset of paired digital-film images. As our image dataset is relatively small, we use a patched training approach, where we randomly sample patches of size $256 \times 256$ from the images before feeding them into the model. We train our model on 400 patches per image in the dataset using the Adam optimizer with a learning rate of $1e-3$ and a batch size of 1. Our experiments are divided into two parts: single-image experiments and full dataset experiments.

For our single-image experiments, we handpicked an image from our dataset that contains a white wall with a light source. It was chosen for the clear blue tint of the film effect and the clearly perceptible grain on the uniform background. For these experiments, we train and evaluate the model on the same image. We first investigate how each loss produces the desired colour effect to gain a better understanding of the interplay between the loss functions and what the model learns.

For the most promising candidates from the single-image experiments, we then
move on to training on the full dataset. The full dataset is split into a
training, validation and test set using a 70-20-10 split ratio, but final evaluation is done on the full dataset.

We also investigate whether adding noise as a 4th input channel to the model improves the its ability to learn the desired visual effects for both single image and full dataset experiments. This is done by concatenating a random uniform noise \footnote{We tried using other forms of noise such as Guassian noise but found that this made little difference.} channel to the input images.  Furthermore, we observed that patched training is very stochastic as each patch only captures a very small portion of the original image and can be very different from the rest of the image. To contain more information in each patch we also experiment applying a resized crop where we first crop the image to a larger patch and then resize to the desired input size. This way we can ensure that each patch contains more information from the original image.

\subsection{Evaluation}
\label{subsec:evaluation}

To evaluate the generated film images, we conduct both qualitative and
quantitative assessments. For quantitative evaluation, we use four image quality and image similarity metrics. We include a reference baseline comparison which is calculated using the original digital image as the prediction.

We use two standard algorithmic metrics, \textbf{Peak Signal-to-Noise Ratio (PSNR)}, which measures the ratio of the maximum possible signal power to the power of corrupting noise, and \textbf{Structural Similarity Index (SSIM)}, which compares structural, luminance, and contrast elements between an original and a reconstructed image in a range from -1 to 1. We also use two learned distance metrics, \textbf{Learned Perceptual Image Patch Similarity (LPIPS)} \cite{LPIPS} and \textbf{PieApp} \cite{PieAPP}, where lower indicates higher similarity, which are trained on human-annotated data to align with human perception of image quality.

\section{Results}
\label{sec:results}

A comprehensive list of all results and experiments can be found in the Appendix Section \ref{sec:all-results}.

\subsection{Single Image Results}

\begin{figure}
    \begin{subfigure}[t]{.19\textwidth}
        \centering
        \includegraphics[width=\linewidth]{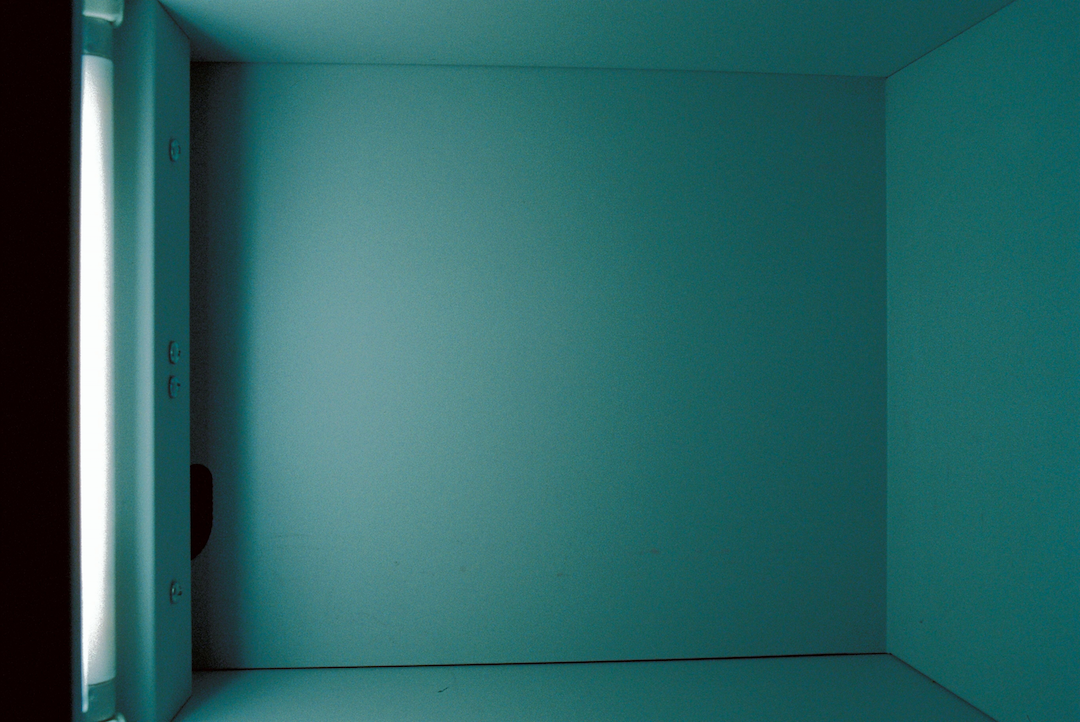}
        \caption{Film \\ Ground truth}
      \end{subfigure}
    \hfill
    \begin{subfigure}[t]{.19\textwidth}
        \centering
        \includegraphics[width=\linewidth]{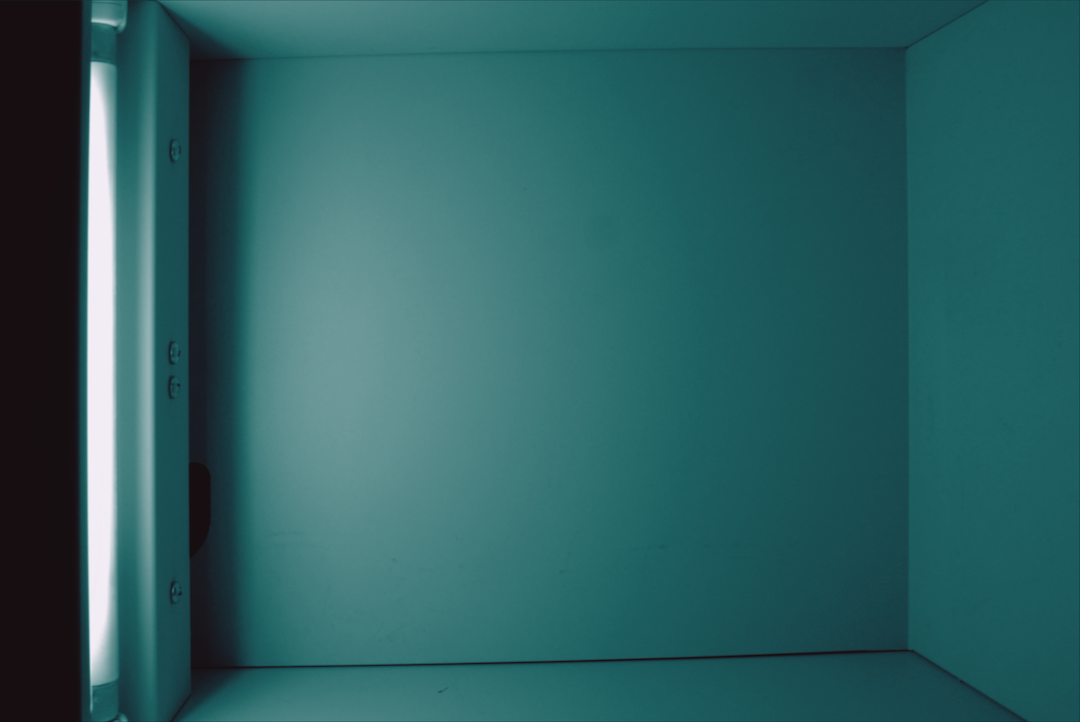}
        \caption{MSE/VGG \\ no noise, resize \\ LPIPS: 0.35 \\ SSIM: \textbf{0.71}}
      \end{subfigure}
    \hfill
    \begin{subfigure}[t]{.19\textwidth}
        \centering
        \includegraphics[width=\linewidth]{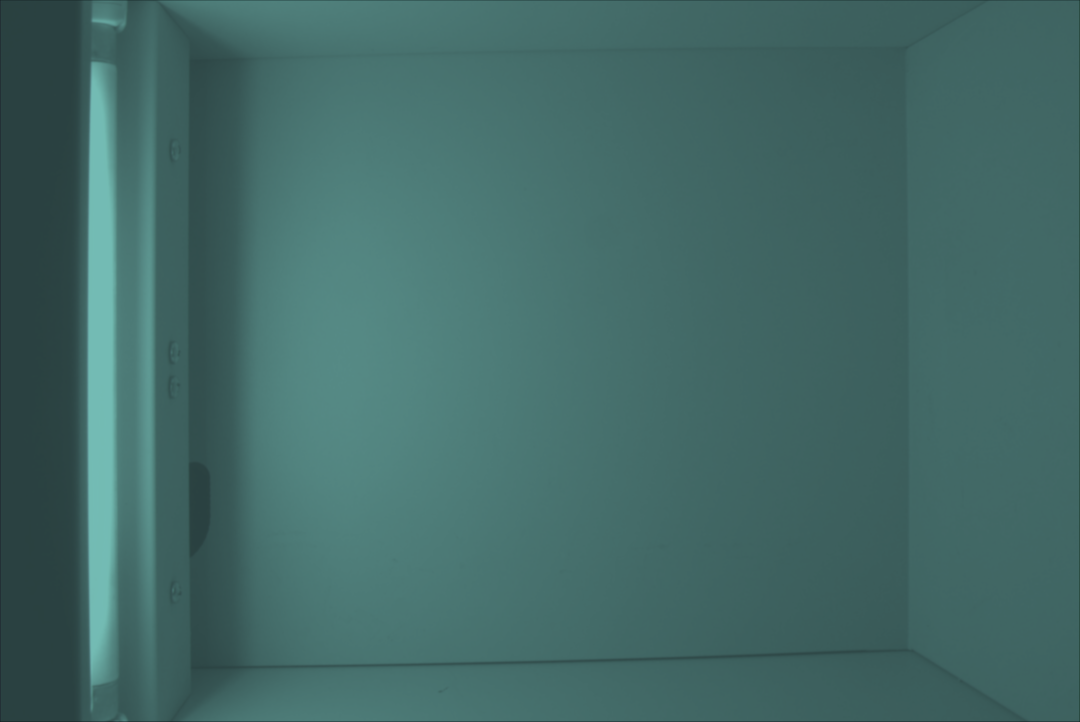}
        \caption{MSE \\ no noise, no resize \\ LPIPS: 0.47 \\ SSIM: 0.67}
      \end{subfigure}
    \hfill
    \begin{subfigure}[t]{.19\textwidth}
      \centering
      \includegraphics[width=\linewidth]{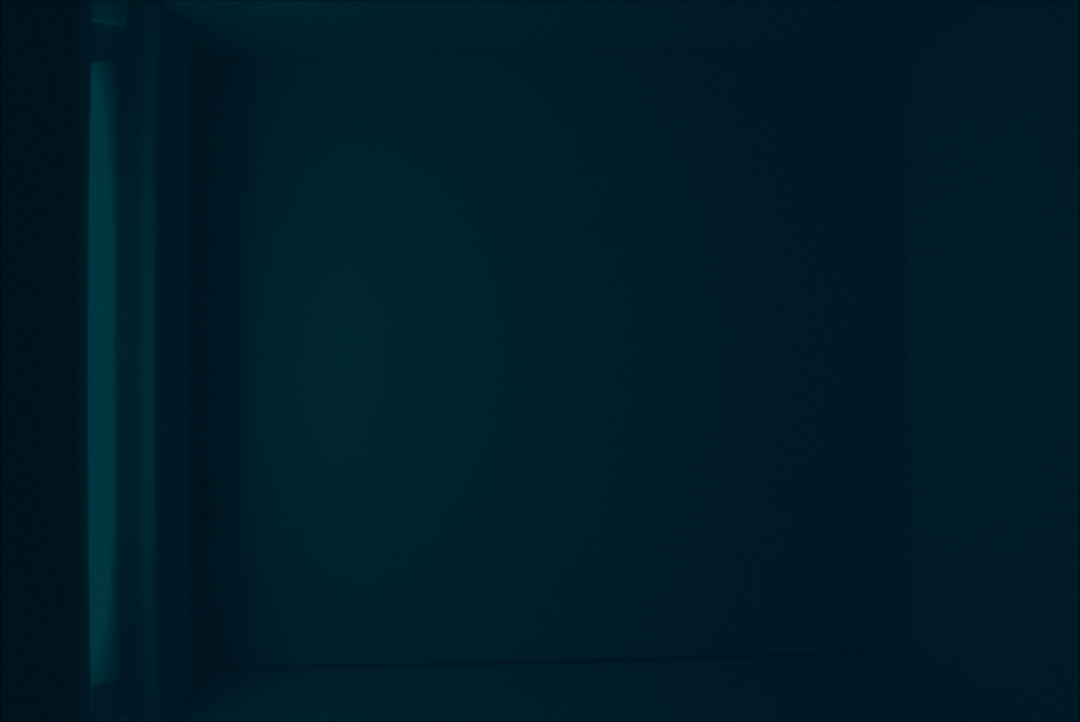}
        \caption{TV-Rel \\ noise, no resize \\ LPIPS: \textbf{0.27} \\ SSIM: 0.27}
    \end{subfigure}
    \hfill
    \begin{subfigure}[t]{.19\textwidth}
        \centering
        \includegraphics[width=\linewidth]{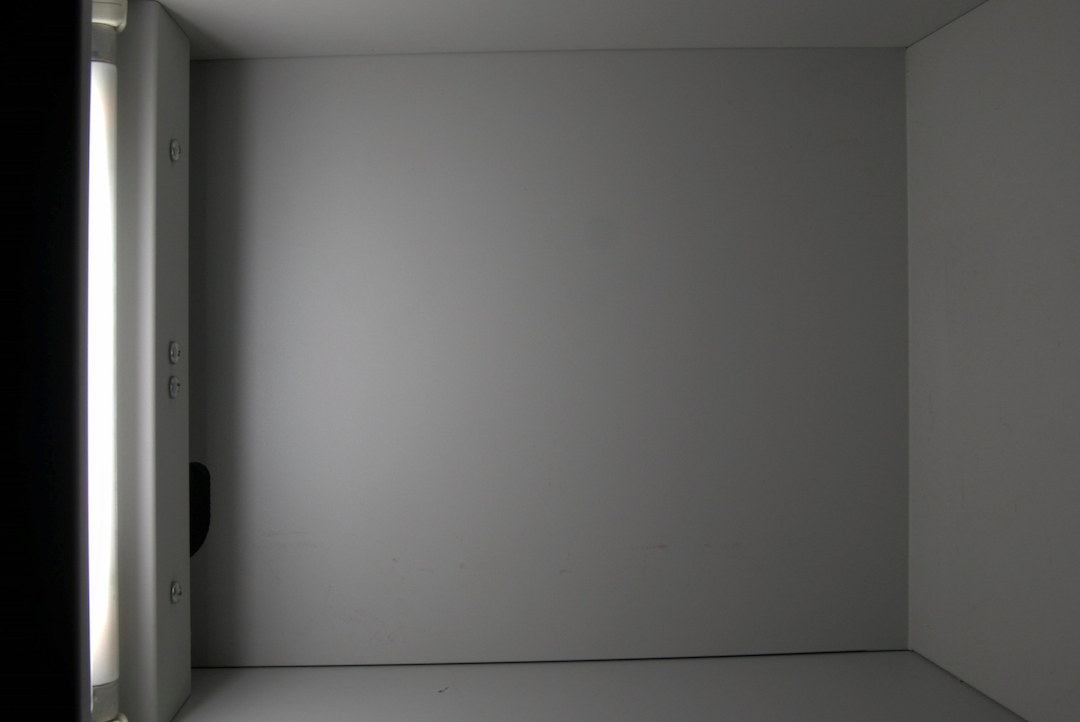}
        \caption{Baseline \\ \\ LPIPS: 0.36 \\ SSIM: 0.64}
      \end{subfigure}
  
    \caption{\textbf{Single Image Select Samples.} Outputs from select loss functions and configurations of noise and resizing. We see that the best performing model is MSE/VGG with resizing, which produces the best colour effect. We also see that LPIPS scores are inconsistent with SSIM and with perception as we can intuitively see that MSE-VGG and MSE are better predicitions than the baseline and TV-Rel.}

    \label{fig:single-image-samples}
\end{figure}

Table \ref{tab:single-image-losses} shows each loss configuration with and without an added noise channel and without resizing. We find that the best performing loss configuration is a combined loss of Color/VGG/TV-Rel, which produces the best SSIM score without noise, and the best PieAPP score in either setting. Notably, the combined losses all perform well, as they account for both the colour and grain effects. The colour-focused losses (MAE, MSE and Colour) all perform well alone, whereas all feature-based losses (VGG and TV-Rel) almost always perform below the baseline. Interestingly, LPIPS scores are inconsistent with the other metrics with the best performing configuration as the baseline and TV-Rel alone with and without noise respectively. This is contrary to what we would be expect as can be seen in a sample of results in Figure \ref{fig:single-image-samples}.

\begin{table}
    \centering
    \captionsetup{justification=centering}
    \caption{
        \textbf{Single Image Losses.} \\Results for each loss configuration with and without noise and no resizing.
    }
    \setlength{\tabcolsep}{0.3em}
    \begin{tabular}{l|cccc|cccc}
        \toprule
        \multicolumn{1}{c}{\textbf{Loss}}& \multicolumn{4}{c}{\textbf{Without Noise}} & \multicolumn{4}{c}{\textbf{With Noise}}\\
        \multicolumn{1}{c}{} & SSIM & PSNR & LPIPS & \multicolumn{1}{c}{PieAPP} & SSIM & PSNR & LPIPS & PieAPP \\
        \midrule
        Color/VGG/TV-Rel & \textbf{0.68} & 19.67 & 0.45 & \textbf{1.42} & 0.67 & 18.01 & 0.37 & \textbf{1.14} \\
        Color/VGG & \textbf{0.68} & \textbf{20.07} & 0.45 & 1.61 & \textbf{0.68} & 17.94 & 0.36 & 1.58 \\
        MSE/VGG & 0.66 & 18.26 & 0.48 & 2.53 & 0.67 & \textbf{19.79} & 0.34 & 2.05 \\
        MAE & 0.65 & 18.61 & 0.50 & 2.12 & \textbf{0.68} & 19.04 & 0.33 & 1.96 \\
        Color & 0.65 & 18.45 & 0.50 & 2.99 & 0.63 & 15.66 & 0.43 & 2.50 \\
        MSE & 0.65 & 18.29 & 0.50 & 2.42 & 0.67 & 19.26 & 0.32 & 1.89 \\
        Baseline & 0.64 & 16.97 & \textbf{0.36} & 3.79 & 0.64 & 16.97 & 0.36 & 3.79 \\
        \midrule
        TV-Rel & 0.17 & 9.84 & 0.60 & 4.84 & 0.27 & 11.11 & \textbf{0.27} & 2.20 \\
        VGG & 0.16 & 9.10 & 0.62 & 4.50 & 0.06 & 8.71 & 0.64 & 5.11 \\
        \bottomrule
    \end{tabular}
    
    \label{tab:single-image-losses}
\end{table}

\begin{figure}

    \begin{subfigure}[t]{.24\textwidth}
        \centering
        \includegraphics[width=\linewidth]{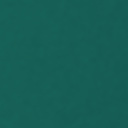}
        
      \end{subfigure}
    \hfill
    \begin{subfigure}[t]{.24\textwidth}
        \centering
        \includegraphics[width=\linewidth]{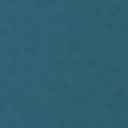}
    \end{subfigure}
    \hfill
    \begin{subfigure}[t]{.24\textwidth}
        \centering
        \includegraphics[width=\linewidth]{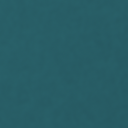}
    \end{subfigure}
    \hfill
    \begin{subfigure}[t]{.24\textwidth}
        \centering
        \includegraphics[width=\linewidth]{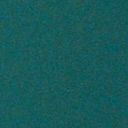}
      \end{subfigure}

    \begin{subfigure}[t]{.24\textwidth}
        \centering
        \includegraphics[width=\linewidth]{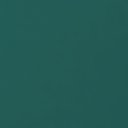}
        \captionsetup{justification=centering}
        \caption{Color/VGG/TV-Rel}
      \end{subfigure}
        \hfill
    \begin{subfigure}[t]{.24\textwidth}
        \centering
        \includegraphics[width=\linewidth]{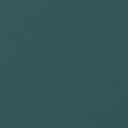}
        \caption{\\MSE/VGG}
      \end{subfigure}
    \hfill
    \begin{subfigure}[t]{.24\textwidth}
        \centering
        \includegraphics[width=\linewidth]{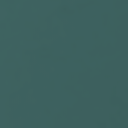}
        \caption{\\MSE}
    \end{subfigure}
    \hfill
    \begin{subfigure}[t]{.24\textwidth}
        \centering
        \includegraphics[width=\linewidth]{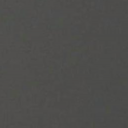}
        \captionsetup{justification=centering}
        \caption{Film (Above) \\ Digital (Below)}
      \end{subfigure}
   
      \centering

    \caption{\textbf{Single Image Noise Comparison.} Outputs from select models without noise (above) and with noise (below), no resizing. We see that when we feed noise into the model, the model learns to produce some variation, especially when the loss contains a feature-based metric like VGG. However, the grain is far from the desired effect.}
    \label{fig:single-image-noise-no-noise}
\end{figure}

The addition of noise has mixed effects. The perceptual metrics significantly reward the addition of noise in almost all cases. SSIM and PSNR react differently depending on the loss: pixel-wise MSE/MAE scores increase slightly, whereas Color loss scores decrease. This can be interpreted as Color loss allowing more noise through the model which is penalised by metrics sensitive to incorrect random noise such as PSNR, where MAE/MSE will even out this noise to be closer to the ground truth, on average. All of this implies that the perceptual metrics better capture the output of noise by the model. Comparisons of outputs from the models with and without noise channels can be seen in Figure \ref{fig:single-image-noise-no-noise}.

We also experiment with resizing cropped patches during training. This is done by taking a crop at a random scale from the image and resizing it to the patch size, $ 256 \times 256$. The results are in Table \ref{tab:single-image-resize}. We show results for only MSE and MSE/VGG as a representative example of how both pixel-wise and combined losses perform. We also experimented with a combination of noise and resizing but found that both performed better alone.  We find that we reach peak SSIM and PSNR scores when resizing the cropped patches, and perceptual metrics also improve across the board.

\begin{figure}[ht]
    \centering

    \begin{subfigure}[b]{0.49\textwidth}
        \centering
        \includegraphics[width=0.49\textwidth]{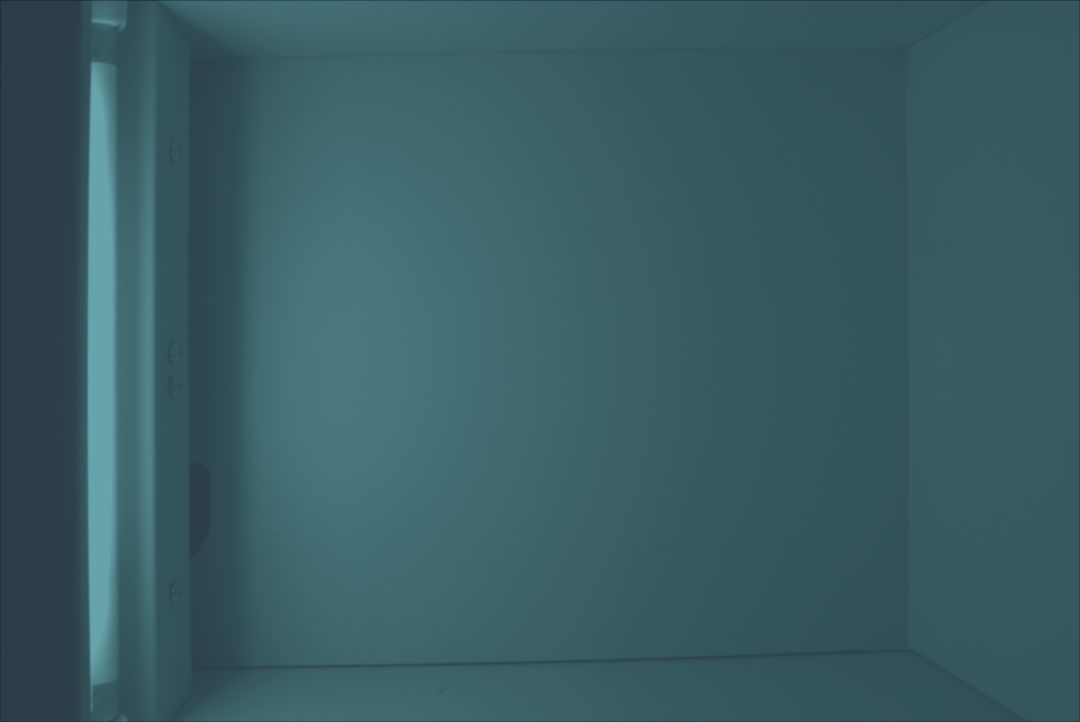}
        \includegraphics[width=0.49\textwidth]{mse-vgg-no-noise-resize-single.png}
        \captionsetup{justification=centering}
        \caption{MSE/VGG, no noise \\ no resize (left) and with resizing (right).}
    \end{subfigure}
    \hfill
    \begin{subfigure}[b]{0.49\textwidth}
        \centering
        \includegraphics[width=0.49\textwidth]{mse-no-noise-no-resize-single.png}
        \includegraphics[width=0.49\textwidth]{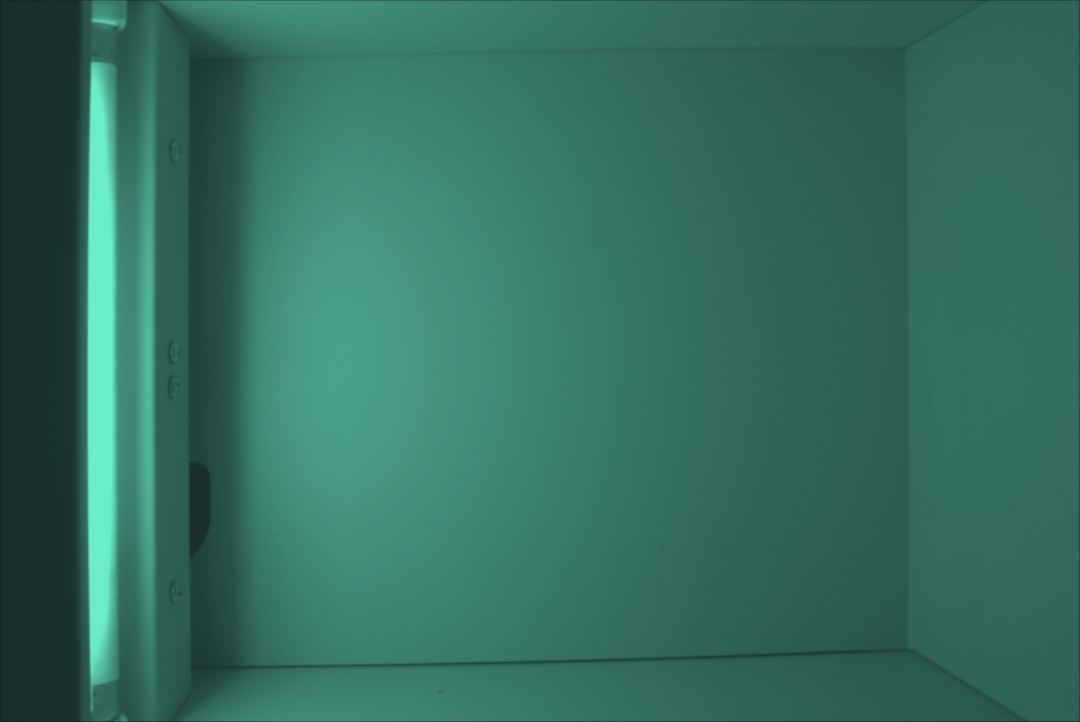}
        \captionsetup{justification=centering}
        \caption{MSE, no noise \\ no resize (left) and with resizing (right).} 
    \end{subfigure}
    \caption{\textbf{Single Image Resizing.} Outputs from MSE and MSE/VGG with and without resizing, and then with resizing and with and without noise. We see that resizing improves the colour effect, but the grain effect is not as strong as in the single image experiments.}
    \label{fig:single-image-resize}
\end{figure}

\begin{figure}[ht]
    \begin{subfigure}[b]{0.49\textwidth}
        \centering
        \includegraphics[width=0.49\textwidth]{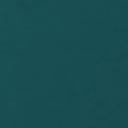}
        \includegraphics[width=0.49\textwidth]{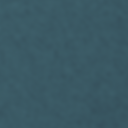}
        \captionsetup{justification=centering}
        \caption{MSE/VGG patch with resizing \\ no noise (left) and with noise (right).} 
    \end{subfigure}
    \hfill
    \begin{subfigure}[b]{0.49\textwidth}
        \centering
        \includegraphics[width=0.49\textwidth]{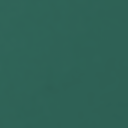}
        \includegraphics[width=0.49\textwidth]{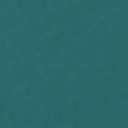}
        \captionsetup{justification=centering}
        \caption{MSE patch with resizing \\ no noise (left) and with noise (right).}
    \end{subfigure}

    \caption{\textbf{Single Image Resizing with Noise.} Outputs from MSE and MSE/VGG with and without noise and resizing. We see that similar grain is still produced even through resizing.}
    \label{fig:single-image-resize-noise}
\end{figure}

We hypothesise that while resizing loses high-resolution information on grain and noise patterns, the model sees a greater variety of areas of the image in a single patch and so may capture the color transformation better which is rewarded by every metric. With the MSE/VGG loss, this colour transformation is a nearly perfect overfit as the outputs in \ref{fig:single-image-resize} show. Furthermore, some grain is still produced when we feed noise into the model despite the resizing operation potentially removing grain, as can be seen in Figure \ref{fig:single-image-resize-noise}.


\begin{table}
    \centering
    \captionsetup{justification=centering}
    \setlength{\tabcolsep}{0.5em}
    \caption{
        \textbf{Single Image Resizing.} \\ Effect of resizing the cropped patches on select losses (no noise).
    }
    \begin{tabular}{lccccc}
        \toprule
        \multicolumn{1}{c}{Loss}& \multicolumn{1}{c}{Resize}&  SSIM & PSNR & LPIPS & PieAPP\\
        \midrule
        MSE/VGG & Yes & \textbf{0.71} & \textbf{23.19} & \textbf{0.35} & \textbf{1.62} \\
        MSE/VGG & No & 0.66 & 18.26 & 0.48 & 2.53 \\
        \midrule
        MSE & Yes & \textbf{0.67} & \textbf{20.03} & \textbf{0.47} & \textbf{1.77} \\
        MSE & No & 0.65 & 18.29 & 0.50 & 2.42 \\

        \bottomrule
    \end{tabular}
    
    \label{tab:single-image-resize}
\end{table}

\subsection{Full Dataset Results}

We select the best performing losses and settings from the single image experiments and evaluate them on the full dataset, with results in Table \ref{tab:full-data-results}.

\begin{table}
    \centering
    \captionsetup{justification=centering}
    \caption{\textbf{Full Dataset Results.} \\ Comparison of best performing losses and configurations on the full dataset.}
    \setlength{\tabcolsep}{0.3em}
    \begin{tabular}{lcccccc}
        \toprule
        Loss & Noise & Resize & SSIM & PSNR & LPIPS & PieAPP \\
        \midrule
        Baseline & - & - & \textbf{0.64} & 21.68 & \textbf{0.26} & 1.96 \\
        MSE/VGG & Yes & Yes & \textbf{0.64} & \textbf{22.87} & 0.27 & \textbf{1.90} \\
        MSE & No & Yes & 0.63 & 22.20 & 0.35 & 2.31 \\
        MAE & No & No & 0.59 & 19.68 & 0.48 & 3.38 \\
        Color/VGG/TV-Rel & Yes & No & 0.59 & 21.42 & 0.41 & 3.07 \\
        MSE/VGG/TV-Rel & Yes & No & 0.58 & 20.43 & 0.46 & 3.54 \\
        \bottomrule
    \end{tabular}
    \label{tab:full-data-results}
\end{table}

\begin{figure}

    \begin{subfigure}[t]{.24\textwidth}
        \centering
        \includegraphics[width=\linewidth]{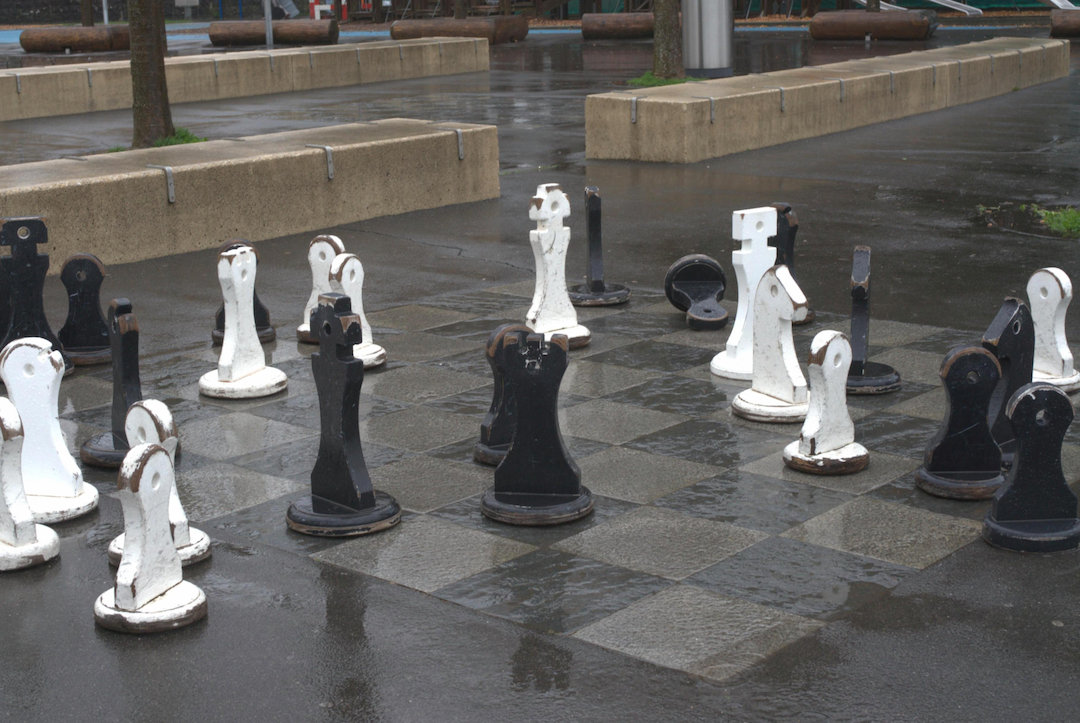}
      \end{subfigure}
    \hfill
    \begin{subfigure}[t]{.24\textwidth}
        \centering
        \includegraphics[width=\linewidth]{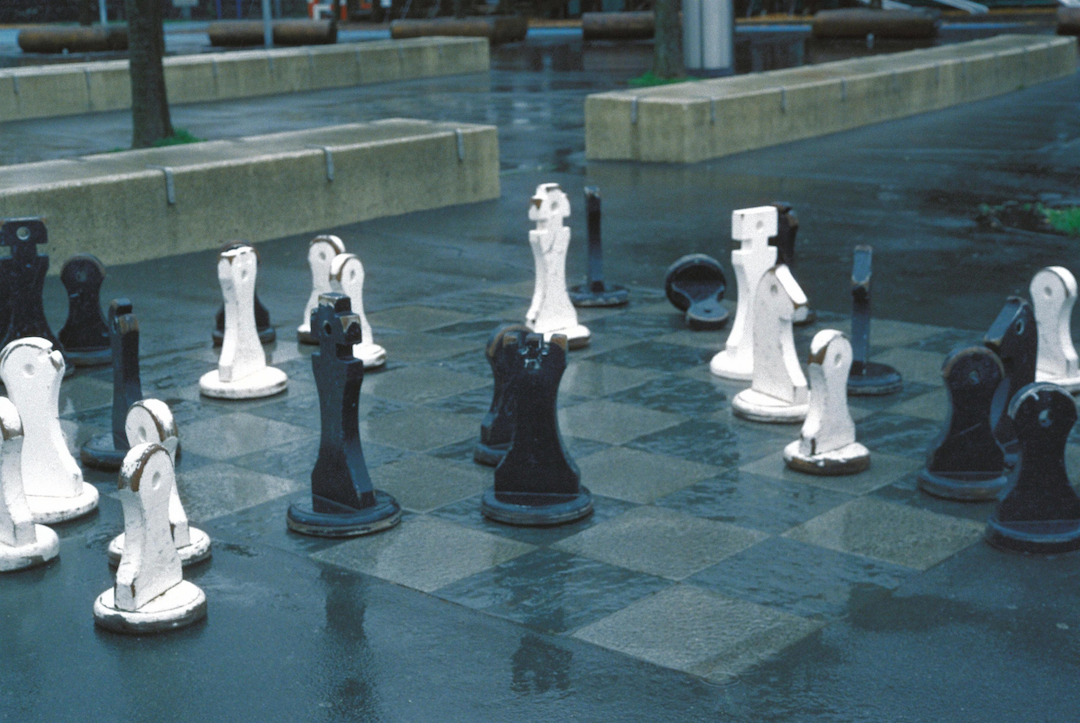}
      \end{subfigure}
    \hfill
    \begin{subfigure}[t]{.24\textwidth}
        \centering
        \includegraphics[width=\linewidth]{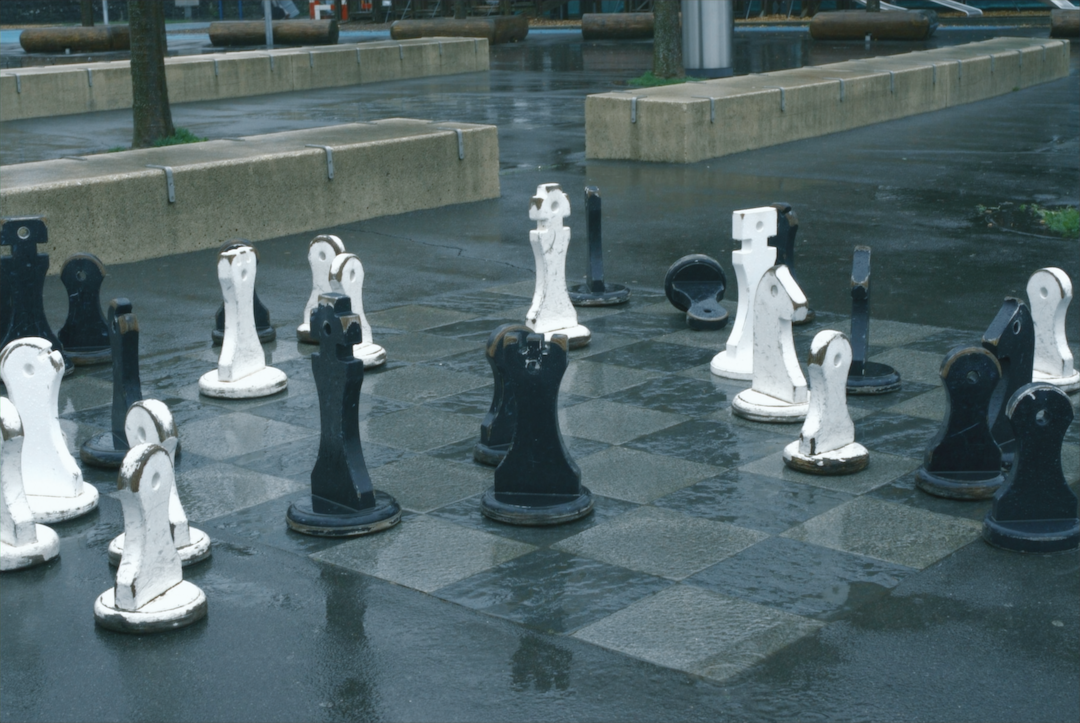}
      \end{subfigure}
    \hfill
    \begin{subfigure}[t]{.24\textwidth}
      \centering
      \includegraphics[width=\linewidth]{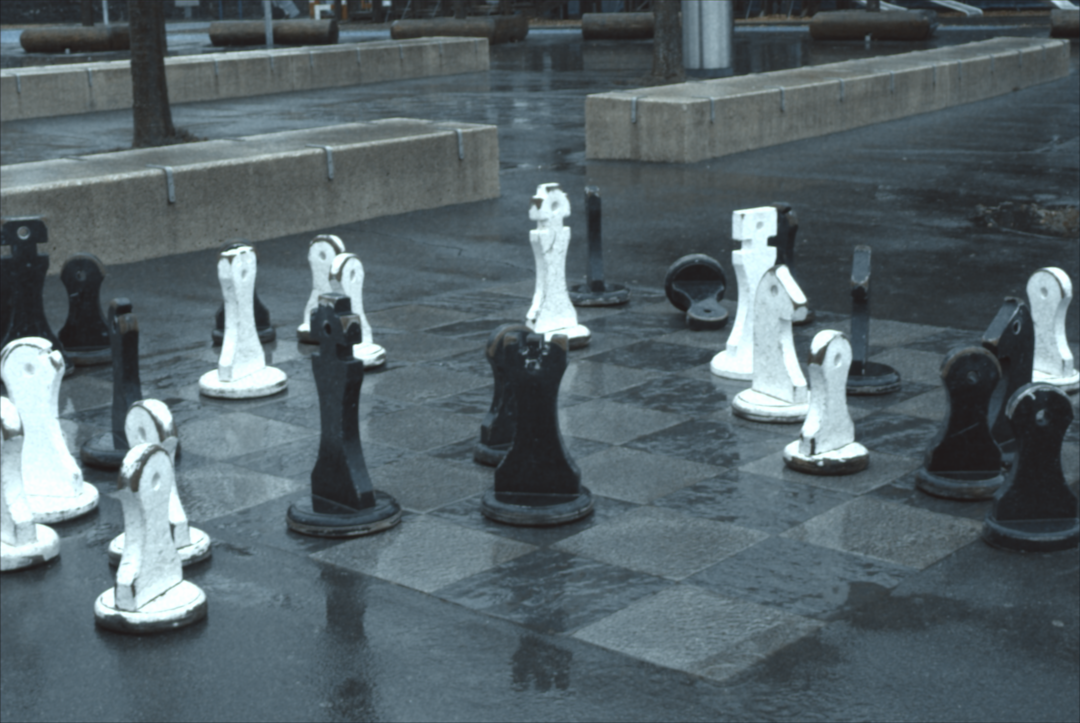}
    \end{subfigure}

    \begin{subfigure}[t]{.24\textwidth}
        \centering
        \includegraphics[width=\linewidth]{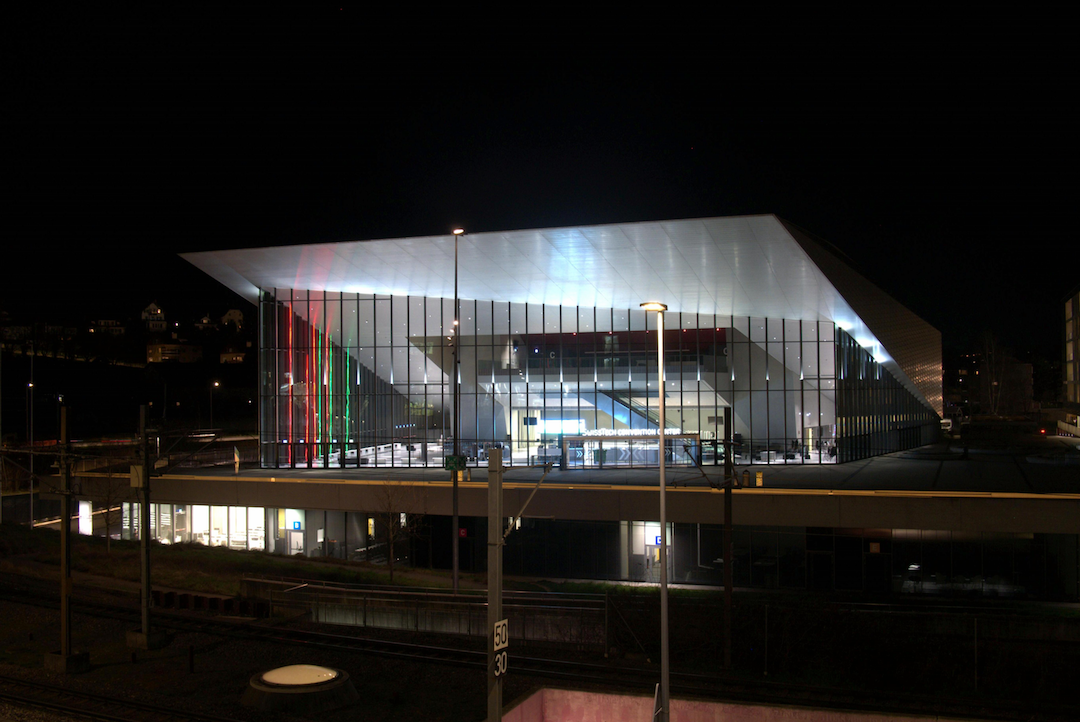}
      \end{subfigure}
    \hfill
    \begin{subfigure}[t]{.24\textwidth}
        \centering
        \includegraphics[width=\linewidth]{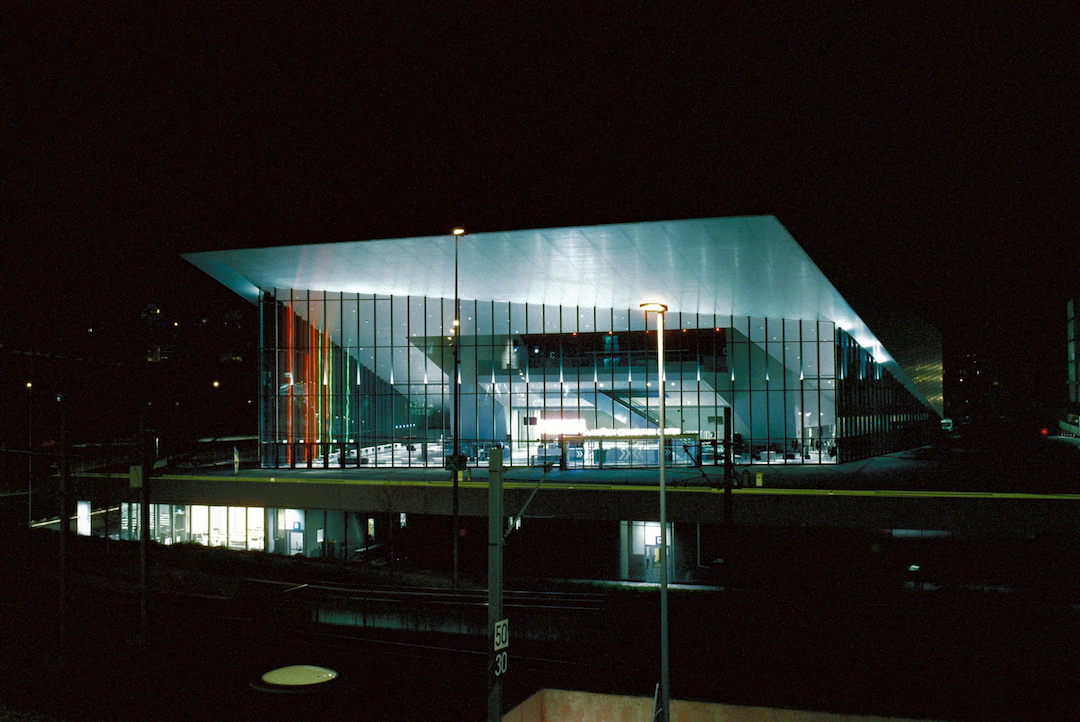}
      \end{subfigure}
    \hfill
    \begin{subfigure}[t]{.24\textwidth}
        \centering
        \includegraphics[width=\linewidth]{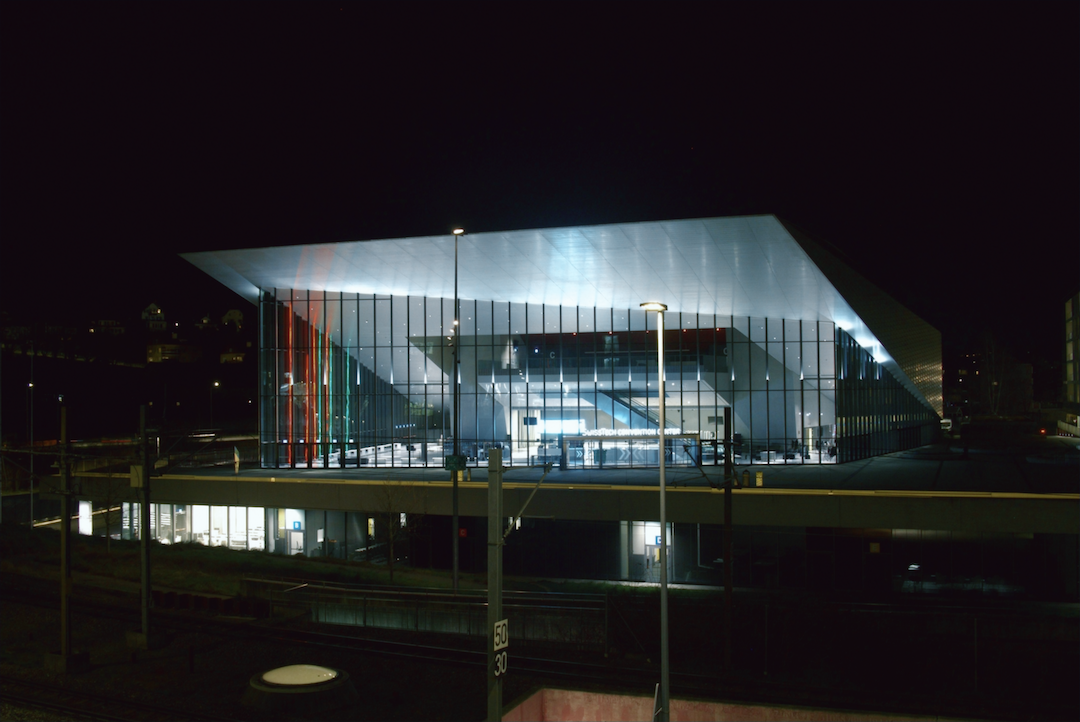}
      \end{subfigure}
    \hfill
    \begin{subfigure}[t]{.24\textwidth}
      \centering
      \includegraphics[width=\linewidth]{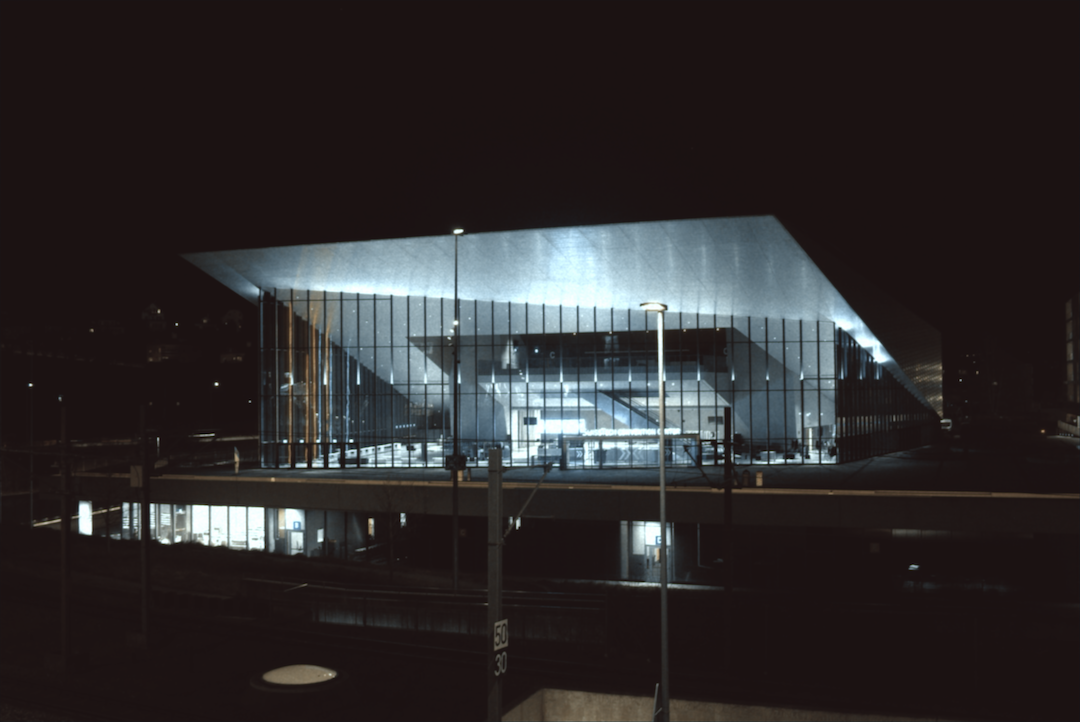}
    \end{subfigure}

    \begin{subfigure}[t]{.24\textwidth}
        \centering
        \includegraphics[width=\linewidth]{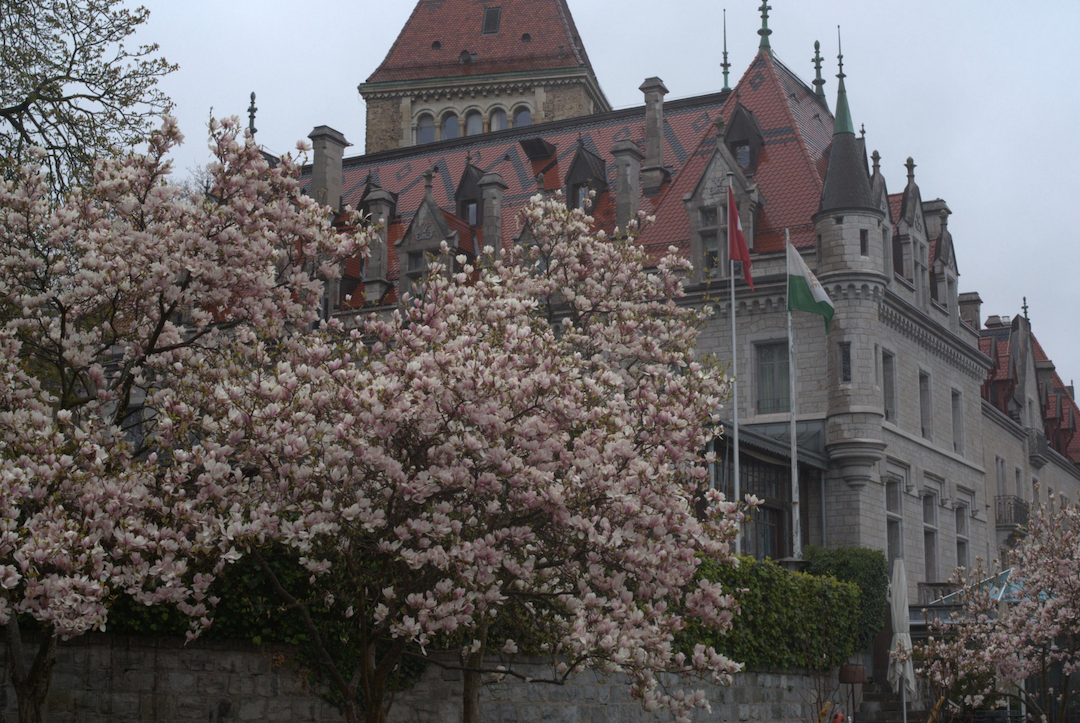}
        \captionsetup{justification=centering}
        \caption{\\Digital}
      \end{subfigure}
    \hfill
    \begin{subfigure}[t]{.24\textwidth}
        \centering
        \includegraphics[width=\linewidth]{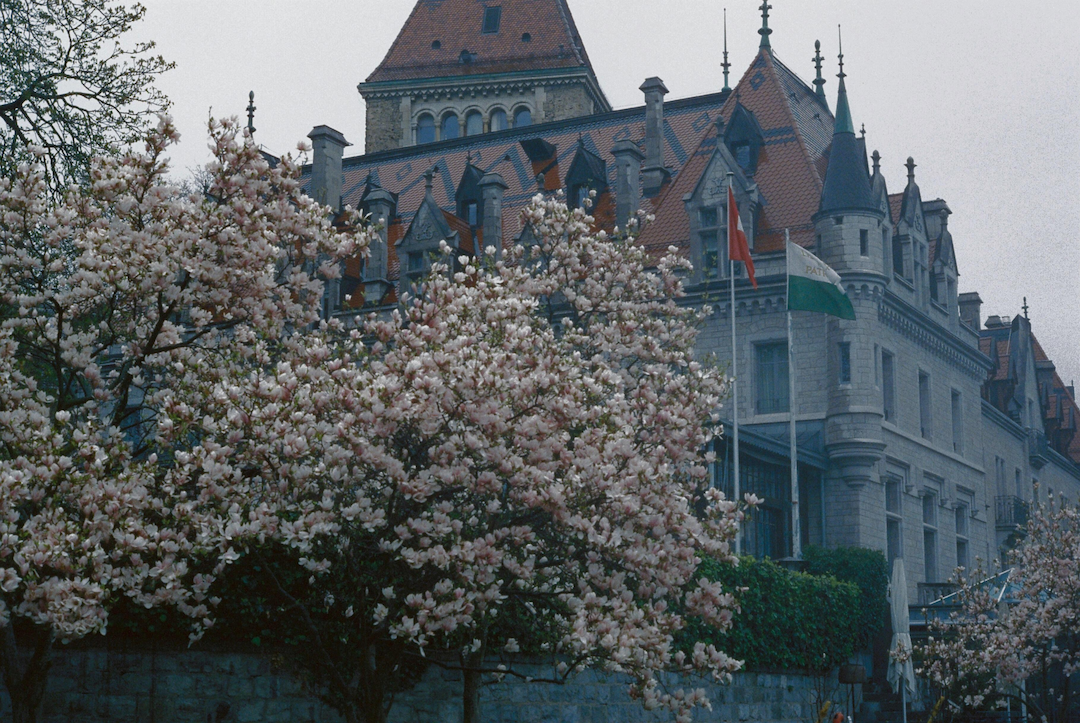}
        \captionsetup{justification=centering}
        \caption{\\Film}
      \end{subfigure}
    \hfill
    \begin{subfigure}[t]{.24\textwidth}
        \centering
        \includegraphics[width=\linewidth]{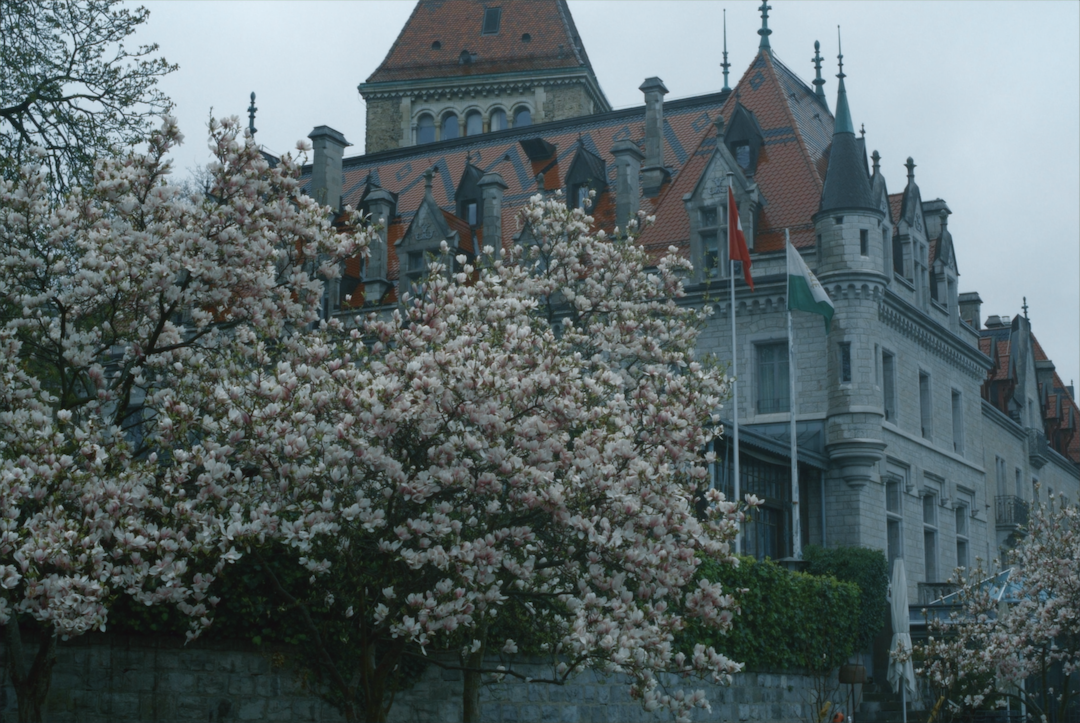}
        \captionsetup{justification=centering}
        \caption{\\MSE/VGG}
      \end{subfigure}
    \hfill
    \begin{subfigure}[t]{.24\textwidth}
      \centering
      \includegraphics[width=\linewidth]{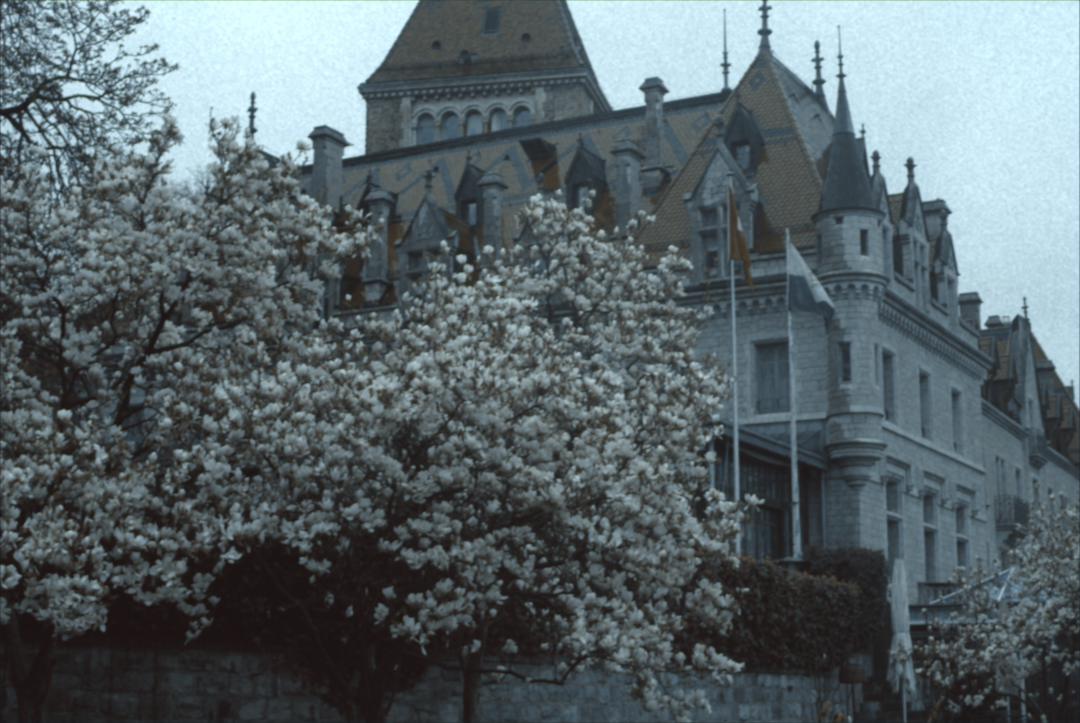}
      \captionsetup{justification=centering}
        \caption{\\Color/VGG/TV-Rel}
    \end{subfigure}

    \caption{\textbf{Full Dataset Select Samples.} Outputs from select models on the full dataset. We see that the best performing model is MSE/VGG with resizing, which produces the best colour effect. We also see that the grain effect is not as strong as in the single image experiments.}
    \label{fig:full-data-results}
\end{figure}

We selected models in terms of high scoring metrics and qualitative evaluation of color and grain production. This leaves us with some models that should produce the best colour: MSE/VGG and MSE with resizing, and other models that should produce the best grain: Color/VGG/TV-Rel, MSE/VGG/TV-Rel with a noise input channel. We find that the best performing model is MSE/VGG with resizing which produces the best metrics across the board. A notable difference compared to the single image results is that our methods now almost never outperform the baseline despite training on the same number of patches per image. This is likely due to the increased complexity of the full dataset, which contains a wider variety of film effects and lighting conditions. The only exception is MSE/VGG which is on par with the baseline. A sample of results can be seen in Figure \ref{fig:full-data-results}.

\begin{figure}
    \begin{subfigure}[t]{.19\textwidth}
        \centering
        \includegraphics[width=\linewidth]{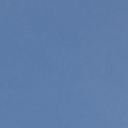}
        \captionsetup{justification=centering}
        \caption{Digital}
      \end{subfigure}
    \hfill
    \begin{subfigure}[t]{.19\textwidth}
        \centering
        \includegraphics[width=\linewidth]{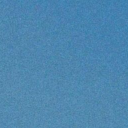}
        \captionsetup{justification=centering}
        \caption{Film}
      \end{subfigure}
    \hfill
    \begin{subfigure}[t]{.19\textwidth}
        \centering
        \includegraphics[width=\linewidth]{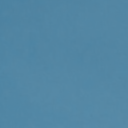}
        \captionsetup{justification=centering}
        \caption{MSE/VGG}
      \end{subfigure}
    \hfill
    \begin{subfigure}[t]{.19\textwidth}
        \centering
        \includegraphics[width=\linewidth]{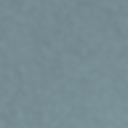}
        \captionsetup{justification=centering}
        \caption{MSE/VGG/TV-Rel}
      \end{subfigure}
      \begin{subfigure}[t]{.19\textwidth}
        \centering
        \includegraphics[width=\linewidth]{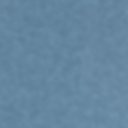}
        \captionsetup{justification=centering}
        \caption{Color/VGG/TV-Rel}
      \end{subfigure}
    \caption{\textbf{Full Dataset Grain Comparison.} Predictions for a patch of sky from select loss functions, all with noise. Similarly to the single image experiments, the models with losses that target some noise produce a more visually pleasing result. The best visual result is produced by Color/VGG/TV-Rel although this is still far from the true grain texture.}
    \label{fig:full-data-grain}
\end{figure}

We continue experimenting with the presence of resizing and noise using MSE and MSE/VGG, with results found in Table \ref{tab:full-data-resize-noise}. Results indicate that once again resizing improves results, but the addition of noise has a less clear effect. The overall best performing model is MSE/VGG with resizing. Noise has little effect. However, the qualitative results tell a different story. In Figure \ref{fig:full-data-grain} we see that the model with noise produces a more visually pleasing result, with a more uniform grain effect. This is not captured by the metrics, which are more sensitive to the colour effect.

\begin{table}
    \centering
    \captionsetup{justification=centering}
    \caption{\textbf{Full Dataset Noise and Resizing.} \\ Comparison of MSE and MSE/VGG with and without noise and resizing.}
    \setlength{\tabcolsep}{0.3em}
    \begin{tabular}{lcccccc}
        \toprule
        Loss & Noise & Resize & SSIM & PSNR & LPIPS & PieAPP \\
        \midrule
        MSE/VGG & Yes & Yes & \textbf{0.64} & \textbf{22.87} & \textbf{0.27} & 1.90 \\
        MSE/VGG & No & Yes & 0.63 & 22.85 & \textbf{0.27} & \textbf{1.84} \\
        MSE/VGG & No & No & 0.58 & 20.47 & 0.50 & 3.67 \\
        \midrule
        MSE & No & Yes & \textbf{0.63} & \textbf{22.20} & \textbf{0.35} & \textbf{2.31} \\
        MSE & Yes & Yes & 0.59 & 20.87 & 0.40 & 3.29 \\
        MSE & No & No & 0.58 & 20.47 & 0.50 & 3.67 \\
        \bottomrule
    \end{tabular}
    \label{tab:full-data-resize-noise}
\end{table}

Comparing results with the related task of enhancement of mobile phone images in \cite{dslr-quality} indicates that our models have a much lower SSIM, but comparable PSNR. They achieve peak SSIM of 0.94 where we observe a peak SSIM of 0.64, whereas peak PSNR is 21.81 and 22.87 respectively. This could be due to a large range of factors such as the the quality and uniformity of the DSLR images, the difference in model architecures or the difference in losses. However, perhaps the biggest reason is the greater complexity of the film effect.


In summary, we find that the best performing model is MSE/VGG with resizing, which produces the best metrics across the board, produces the best colour effect and often the most visually pleasing result too. The addition of noise has mixed effects but tends to help in noise production, and resizing the cropped patches during training significantly improves the learned colour transformation. The perceptual metrics better capture the output of noise by the model, and resizing allows the model to see a greater variety of areas of the image in a single patch. However, the full dataset is more complex and the models do not outperform the baseline as they did in the single image experiments. Our best model learns a good colour mapping, but the grain effect is not as strong as in the single image experiments, nor is it exactly the kind of grain observed.
\section{Conclusion}
\label{sec:conclusion}

In this work, we have explored the use of convolution neural networks in modeling the effect given by Cinestill800T film. We focused on the use of different loss functions as well as the addition of a noise channel to the input, and the use of random scales of patches during training. We find that a combination of MSE/VGG gives the best colour production which is significantly helped by resizing, and that the addition of a noise channel and another loss such as TV-Rel produces some grain.

Our contributions include the creation of a dataset of paired images taken with a film and digital camera, and all of our code in Pytorch, the details of which can be found in Section \ref{sec:code-documentation} in the appendix.

Our results are severely limited by the small size and great variety of our dataset, meaning that the model must learn a complex function with relatively little training data. This is evidenced by the lack of sign of halation ever being produced by our models, as they were simply not trained on enough patches containing halation. We also restricted ourselves to a pure deep learning approach, and did not explore the use of statistical models for film grain, halation, and colour hue.

\section{Further Work}
\label{sec:further-work}

While we tried many combinations of hyperpamaters, we did not rigorously explore all options. Further work could systematically investigate the effect of kernel size, model size, patch size or the weightings for each loss function when combined. The dataset could be also narrowed focused to include only scenes of a similar nature to give the model an easier function to learn. This could be taken to the extreme for halation, where only patches containing some halation could be fed to the model. A natural extension would be to condition the model on features of the image such as dynamic range. 

For grain production, an implicitly defined texture loss that more precisely captures the grain seen in films could be added, such as GCLM \cite{glcm} or notably an adversarial loss as in \cite{dslr-quality}. Avenues we only briefly explored could be tested in more depth. This includes the paired auto-encoder architecture from \cite{raw-to-raw} and the contextual bilateral loss from \cite{zoom-to-learn}, both contained in our code. Lastly, the model could be combined with a non-ML approach to see if a hybrid model can produce better results.

\newpage
\bibliographystyle{plain}
\bibliography{main}

\begin{thebibliography}{10}
\providecommand{\url}[1]{\texttt{#1}}
\providecommand{\urlprefix}{URL }
\providecommand{\doi}[1]{https://doi.org/#1}

\bibitem{raw-to-raw}
Afifi, M., Abuolaim, A.: Semi-supervised raw-to-raw mapping. CoRR
  \textbf{abs/2106.13883} (2021), \url{https://arxiv.org/abs/2106.13883}

\bibitem{opencv}
Bradski, G.: {The OpenCV Library}. Dr. Dobb's Journal of Software Tools  (2000)

\bibitem{deep-colorisation}
Cheng, Z., Yang, Q., Sheng, B.: Deep colorization. CoRR
  \textbf{abs/1605.00075} (2016), \url{http://arxiv.org/abs/1605.00075}

\bibitem{large-scale-colorisation}
Deshpande, A., Rock, J., Forsyth, D.A.: Learning large-scale automatic image
  colorization. 2015 IEEE International Conference on Computer Vision (ICCV)
  pp. 567--575 (2015), \url{https://api.semanticscholar.org/CorpusID:13195504}

\bibitem{image-super-resolution-with-deep-networks}
Dong, C., Loy, C.C., He, K., Tang, X.: Image super-resolution using deep
  convolutional networks (2015)

\bibitem{scaling-painting-style-transfer}
Galerne, B., Raad, L., Lezama, J., Morel, J.M.: Scaling painting style transfer
  (2022)

\bibitem{image-style-transfer}
Gatys, L.A., Ecker, A.S., Bethge, M.: Image style transfer using convolutional
  neural networks. In: 2016 IEEE Conference on Computer Vision and Pattern
  Recognition (CVPR). pp. 2414--2423 (2016). \doi{10.1109/CVPR.2016.265}

\bibitem{arbitrary-style-transfer}
Huang, X., Belongie, S.: Arbitrary style transfer in real-time with adaptive
  instance normalization (2017)

\bibitem{dslr-quality}
Ignatov, A., Kobyshev, N., Vanhoey, K., Timofte, R., Gool, L.V.: Dslr-quality
  photos on mobile devices with deep convolutional networks. CoRR
  \textbf{abs/1704.02470} (2017), \url{http://arxiv.org/abs/1704.02470}

\bibitem{conditionalgan}
Isola, P., Zhu, J.Y., Zhou, T., Efros, A.A.: Image-to-image translation with
  conditional adversarial networks (2018)

\bibitem{resolution-perceptual-losses}
Johnson, J., Alahi, A., Fei-Fei, L.: Perceptual losses for real-time style
  transfer and super-resolution (2016)

\bibitem{perceptual-losses-style-transfer}
Johnson, J., Alahi, A., Fei-Fei, L.: Perceptual losses for real-time style
  transfer and super-resolution (2016)

\bibitem{gan-progressive-stability-resolution}
Karras, T., Aila, T., Laine, S., Lehtinen, J.: Progressive growing of gans for
  improved quality, stability, and variation (2018)

\bibitem{gan-style-based-resolution}
Karras, T., Laine, S., Aila, T.: A style-based generator architecture for
  generative adversarial networks (2019)

\bibitem{accurate-image-super-resolution}
Kim, J., Lee, J.K., Lee, K.M.: Accurate image super-resolution using very deep
  convolutional networks (2016)

\bibitem{gan-photo-realistic-super-resolution}
Ledig, C., Theis, L., Huszar, F., Caballero, J., Cunningham, A., Acosta, A.,
  Aitken, A., Tejani, A., Totz, J., Wang, Z., Shi, W.: Photo-realistic single
  image super-resolution using a generative adversarial network (2017)

\bibitem{intrinsic-colorisation}
Liu, X., Wan, L., Qu, Y., Wong, T.T., Lin, S., Leung, C.S., Heng, P.A.:
  Intrinsic colorization. ACM Trans. Graph.  \textbf{27}(5) (2008).
  \doi{10.1145/1409060.1409105}, \url{https://doi.org/10.1145/1409060.1409105}

\bibitem{deep-photo-style-transfer}
Luan, F., Paris, S., Shechtman, E., Bala, K.: Deep photo style transfer (2017)

\bibitem{contextual-loss}
Mechrez, R., Talmi, I., Zelnik-Manor, L.: The contextual loss for image
  transformation with non-aligned data (2018)

\bibitem{pulse-gan-perceptual-super-resolution}
Menon, S., Damian, A., Hu, S., Ravi, N., Rudin, C.: Pulse: Self-supervised
  photo upsampling via latent space exploration of generative models (2020)

\bibitem{flann}
Muja, M., Lowe, D.: Fast approximate nearest neighbors with automatic algorithm
  configuration. vol.~1, pp. 331--340 (01 2009)

\bibitem{film-grain-rendering}
Newson, A., Faraj, N., Galerne, B., Delon, J.: {Realistic Film Grain
  Rendering}. {Image Processing On Line}  \textbf{7},  165--183 (2017),
  \url{https://doi.org/10.5201/ipol.2017.192}

\bibitem{PieAPP}
Prashnani, E., Cai, H., Mostofi, Y., Sen, P.: Pieapp: Perceptual image-error
  assessment through pairwise preference. CoRR  \textbf{abs/1806.02067} (2018),
  \url{http://arxiv.org/abs/1806.02067}

\bibitem{unet}
Ronneberger, O., Fischer, P., Brox, T.: U-net: Convolutional networks for
  biomedical image segmentation (2015)

\bibitem{orb}
Rublee, E., Rabaud, V., Konolige, K., Bradski, G.: Orb: an efficient
  alternative to sift or surf. pp. 2564--2571 (11 2011).
  \doi{10.1109/ICCV.2011.6126544}

\bibitem{vgg}
Simonyan, K., Zisserman, A.: Very deep convolutional networks for large-scale
  image recognition (2015)

\bibitem{recovering-texture-super-resolution}
Wang, X., Yu, K., Dong, C., Loy, C.C.: Recovering realistic texture in image
  super-resolution by deep spatial feature transform (2018)

\bibitem{esrgan-super-resolution}
Wang, X., Yu, K., Wu, S., Gu, J., Liu, Y., Dong, C., Loy, C.C., Qiao, Y., Tang,
  X.: Esrgan: Enhanced super-resolution generative adversarial networks (2018)

\bibitem{glcm}
Wang, Y.D., Swietojanski, P., Armstrong, R.T., Mostaghimi, P.: Pixel
  co-occurence based loss metrics for super resolution texture recovery (2020),
  \url{https://openreview.net/forum?id=rylrI1HtPr}

\bibitem{LPIPS}
Zhang, R., Isola, P., Efros, A.A., Shechtman, E., Wang, O.: The unreasonable
  effectiveness of deep features as a perceptual metric. CoRR
  \textbf{abs/1801.03924} (2018), \url{http://arxiv.org/abs/1801.03924}

\bibitem{zoom-to-learn}
Zhang, X.C., Chen, Q., Ng, R., Koltun, V.: Zoom to learn, learn to zoom. CoRR
  \textbf{abs/1905.05169} (2019), \url{http://arxiv.org/abs/1905.05169}

\bibitem{cyclicgan}
Zhu, J.Y., Park, T., Isola, P., Efros, A.A.: Unpaired image-to-image
  translation using cycle-consistent adversarial networks (2020)

\end{thebibliography}

\newpage
\section{Appendix}
\label{sec:appendix}

\subsection{Code Documentation}
\label{sec:code-documentation}

The code for this project is available at \url{https://github.com/mikasenghaas/sillystill/tree/main}. The code is written in Python and uses \href{https://pytorch.org/}{PyTorch}, \href{https://lightning.ai/docs/pytorch/stable/}{Lightning} and \href{https://hydra.cc/}{Hydra} for configuration. The code is structured as follows:

\begin{itemize}
    \item \texttt{/src} contains the main code for the project, including the data processing, models, losses and training scripts.
        \begin{itemize}
            \item \texttt{/data} contains all datasets and Lightning datamodules. 
            \item \texttt{/models} contains the PyTorch model classes and loss functions.
            \item \texttt{/eval} contains wrapper classes for evaluation metrics.
            \item \texttt{/utils} contains a variety of utility functions.
            \item \texttt{train.py} is the main training script.
            \item \texttt{dev.py} is a developer script for testing code without using Lightning or Hydra.
            \item \texttt{preprocess.py} is a script for preprocessing the dataset.
            \item \texttt{evaluate.py} is an evaluation script.
            \item \texttt{unsplash.py} is a script for downloading images from Unsplash.
        \end{itemize}
    \item \texttt{/data} contains our dataset
    \item \texttt{/configs} contains the Hydra configuration files for the experiments.
    \item \texttt{/notebooks} contains Jupyter notebooks for investigating the data processing, losses, models and evaluation.
    \item \texttt{/report} contains the Latex files for the report.
\end{itemize}

In order to train a model, the \texttt{src/train.py} script can be run with the desired configuration. You can specify a hydra configuration with flags or by editing the configuration files. The two experiments in this report can be run with the following commands:

\begin{verbatim}
python src/train.py experiment=single-image logger=none
python src/train.py experiment=full-data logger=none
\end{verbatim}

To edit the loss functions or other parameters, you can edit the configuration file \texttt{configs/experiment/single-image.yaml}. The model will be trained, and inference will be run automatically, with results logged locally. If you want to log  to weights and biases, you can remove the logger flag.

\newpage
\subsection{All Results}
\label{sec:all-results}

\begin{table}[ht]
    \scriptsize
    \centering
    \caption{
        \textbf{Single Image.} All Results}
    \begin{tabular}{lcc|cccc}
        \toprule
        \multicolumn{3}{c}{\textbf{Configuration}}& \multicolumn{4}{c}{\textbf{Evaluation}} \\
        Loss & Noise & Resized & SSIM & PSNR & LPIPS & PieAPP \\
        \midrule
        MSE + VGG & No & Yes & \textbf{0.71} & \textbf{23.19} & 0.35 & 1.62 \\
        MAE & Yes & No & 0.68 & 19.04 & 0.33 & 1.96 \\
        Color + VGG + TV-Rel & No & No & 0.68 & 19.67 & 0.45 & 1.42 \\
        Color + VGG & No & No & 0.68 & 20.07 & 0.45 & 1.61 \\
        Color + VGG & Yes & No & 0.68 & 17.94 & 0.36 & 1.58 \\
        Color + VGG + TV-Rel & Yes & No  & 0.67 & 18.01 & 0.37 & \textbf{1.14} \\
        MSE & No & Yes & 0.67 & 20.03 & 0.47 & 1.77 \\
        MSE & Yes & No & 0.67 & 19.26 & 0.32 & 1.89 \\
        MSE + VGG & Yes & No & 0.67 & 19.79 & 0.34 & 2.05 \\
        MSE & Yes & Yes & 0.67 & 18.61 & 0.30 & 1.87 \\
        MSE + TV-Rel + VGG & Yes & No & 0.66 & 20.14 & 0.30 & 1.84 \\
        MSE + VGG & No & No & 0.66 & 18.26 & 0.48 & 2.53 \\
        MAE & No & No & 0.65 & 18.61 & 0.50 & 2.12 \\
        Color & No & No & 0.65 & 18.45 & 0.50 & 2.99 \\
        MSE & No & No & 0.65 & 18.29 & 0.50 & 2.42 \\
        \textbf{Baseline} & - & - & 0.64 & 16.97 & 0.36 & 3.79 \\
        \midrule
        Color & Yes & No & 0.63 & 15.66 & 0.43 & 2.50 \\
        TV-Rel & Yes & No & 0.27 & 11.11 & \textbf{0.27} & 2.20 \\
        TV-Rel & No & No & 0.17 & 9.84 & 0.60 & 4.84 \\
        VGG & No & No & 0.16 & 9.10 & 0.62 & 4.50 \\
        VGG & Yes & No & 0.06 & 8.71 & 0.64 & 5.11 \\
        \bottomrule
    \end{tabular}
    \label{tab:all-single-image-results}
\end{table}

\begin{table}[ht]
    \scriptsize
    \centering
    \setlength{\tabcolsep}{0.3em}
    \caption{\textbf{Full Dataset.} All results}
    \begin{tabular}{lcc|cccc}
        \toprule
        \multicolumn{3}{c}{\textbf{Configuration}} & \multicolumn{4}{c}{\textbf{Evaluation}} \\
        Loss & Resize & Noise & SSIM & PSNR & LPIPS & PieAPP \\
        \midrule
        \textbf{Baseline} & - & - & \textbf{0.64} & 21.68 & \textbf{0.26} & 1.96 \\
        \hline
        MSE + VGG & Yes & Yes & \textbf{0.64} & \textbf{22.87} & 0.27 & 1.90 \\
        MSE + VGG & Yes & No & 0.63 & 22.85 & 0.27 & \textbf{1.84} \\
        MSE & Yes & No & 0.63 & 22.20 & 0.35 & 2.31 \\
        MSE & Yes & No & 0.59 & 19.68 & 0.48 & 3.38 \\
        Color + VGG + TV-Rel & No & Yes & 0.59 & 21.42 & 0.41 & 3.07 \\
        MSE & No & No & 0.58 & 20.47 & 0.50 & 3.67 \\
        MSE + VGG & No & No & 0.58 & 20.47 & 0.50 & 3.67 \\
        MSE + VGG + TV-Rel & No & Yes & 0.58 & 20.43 & 0.46 & 3.54 \\
        \bottomrule
    \end{tabular}
    \label{tab:all-full-dataset-results}
\end{table}

\end{document}